\documentclass[twoside,leqno,twocolumn]{article}

\usepackage[letterpaper]{geometry}

\usepackage{ltexpprt}
\usepackage{color,xcolor}
\definecolor{mydarkblue}{rgb}{0,0.08,0.45}
\usepackage[colorlinks=true,
    linkcolor=mydarkblue,
    citecolor=mydarkblue,
    filecolor=mydarkblue,
    urlcolor=mydarkblue]{hyperref}
\usepackage[numbers, compress]{natbib}
\usepackage[page]{appendix}
\usepackage{url}
\usepackage{mathtools}
\usepackage{subfig}
\usepackage{booktabs}
\usepackage{float}
\usepackage{amsmath,amsfonts,amssymb}
\DeclareMathOperator*{\argmin}{arg\,min}
\newcommand*\samethanks[1][\value{footnote}]{\footnotemark[#1]}

\makeatletter
\newcommand\notsotiny{\@setfontsize\notsotiny{6.5}{8.0}}
\makeatother

\begin{document}

\newcommand\relatedversion{}
\renewcommand\relatedversion{\thanks{The full version of the paper can be accessed at \protect\url{https://arxiv.org/abs/1910.08527}}} 

\title{\Large Masked Gradient-Based Causal Structure Learning}
\author{\hspace{1.5em}Ignavier Ng\thanks{Equal contribution.} \thanks{University of Toronto.}
\and Shengyu Zhu\samethanks[1] \thanks{Huawei Noah's Ark Lab. Email: zhushengyu@huawei.com.}
\and Zhuangyan Fang\samethanks[1] \thanks{Peking University.}
\and Haoyang Li\thanks{École Polytechnique.}\hspace{1.5em}
\and Zhitang Chen\samethanks[3]
\and Jun Wang\thanks{University College London.}
}

\date{}

\maketitle


\fancyfoot[R]{\scriptsize{Copyright \textcopyright\ 2022 by SIAM\\
Unauthorized reproduction of this article is prohibited}}





\begin{abstract} \small This paper studies the problem of learning causal structures from observational data. We  reformulate the Structural Equation Model (SEM) with additive noises in a form parameterized by binary graph adjacency matrix and show that, if the original SEM is identifiable, then the binary adjacency matrix can be identified up to super-graphs of the true causal graph under mild conditions. We then utilize the reformulated SEM to develop a causal structure learning method that can be efficiently trained using gradient-based optimization, by leveraging a smooth characterization on acyclicity and the Gumbel-Softmax approach to approximate the binary adjacency matrix. It is found that the obtained entries are typically near zero or one and can be easily thresholded to identify the edges. We conduct experiments on synthetic and real datasets to validate the effectiveness of the proposed method, and show that it readily includes different smooth model functions and achieves a much improved performance on most datasets considered.

\end{abstract}
\section{Introduction}\label{intro}
Causal graphical models defined on Directed Acyclic Graphs (DAGs) find applications in many sciences including economics \citep{Koller2009} and biology \citep{pearl2009causality,Sachs2005causal}. Although controlled experiments can discover the structures effectively, they are often expensive or practically impossible. Learning causal structures from observational data is hence appealing and has been made possible under proper conditions \cite{spirtes2000causation,pearl2009causality,Peters2017book}.

Existing approaches for structure learning roughly fall into two classes: constraint- and score-based methods. Constraint-based methods, such as PC and fast causal inference \cite{spirtes2000causation}, use conditional independence tests to find initial skeleton and then determine the edge directions according to certain orientation rules~\citep{meek1995casual,zhang2008completeness}. Score-based methods, on the other hand, evaluate the candidate graphs with a predefined score function and search for the ones with the optimal score. Due to the combinatorial nature of acyclicity constraint \citep{Chickering1996learning}, exact search algorithms like dynamic programming \cite{Koivisto2004exact,Singh2005finding} only work for small problems and therefore local greedy strategies are often adopted to perform the search, such as Greedy Equivalence Search (GES) \citep{chickeringo2002optimal}. We refer the reader to \citet{Glymour2019review,Spirtes2018search} for more details and also a review of other methods.

More recently, \citet{zheng2018dags} have proposed NOTEARS, a score-based method that formulates the structure learning problem of linear Structural Equation Models (SEMs) as a continuous one using a smooth characterization of acyclicity. Subsequent works such as DAG-GNN \citep{Yu19DAGGNN} and GraN-DAG \citep{Lachapelle2019grandag} have extended it to handle nonlinear cases. With smooth score functions, these methods utilize gradient-based optimization to learn DAGs by estimating some {\it weighted} graph adjacency matrices. NOTEARS and DAG-GNN assume specific forms of SEMs where weighted adjacency matrices naturally exist; their performance usually degrades when the data model does not follow these forms. GraN-DAG uses path products to construct an equivalent weighted matrix representing the estimated graph, which generally requires much effort to obtain explicitly given a particular model function. Notice that choosing a right model function to fit the underlying relationships plays an important role in these methods, since the adopted score functions rely on reconstructed observations. For example, it was shown in \citet{Zhu2020causal} that GraN-DAG performed poorly on linear SEMs but could be much improved by modifying it with linear model functions. In addition, numeric methods usually result in a number of estimated entries in a contiguous region near zero; it becomes key to picking a proper threshold to obtain exact DAGs while keeping most, if not all, true positives, but current gradient-based approaches mostly rely on prior knowledge or human experience.

In this work, we develop a gradient-based optimization framework for structure learning, called \emph{Masked gradient-based Causal Structure Learning} (MCSL), which (1) flexibly includes different model functions (e.g., neural networks, polynomial functions) with little extra effort, and (2) is easily thresholded to identify the edges. To achieve these, we reformulate the SEM with additive noises in a form parameterized by {\emph{binary}} adjacency matrix that exists for any data model and is called \emph{mask} in the paper---the binary valued entry can be used to remove the effect of a variable on the output of any function.
We characterize the structure identification issue and show that if the original SEM is identifiable, the binary adjacency matrix can be identified as super-graphs of the true causal graph.
To enable an efficient learning of binary valued entries, we leverage the Gumbel-Softmax approach and develop a gradient-based structure learning method with carefully devised training procedure. The resulting entries of the estimate are mostly near either zero or one, so that the edges are easily identified by thresholding at $0.5$.
We conduct experiments on synthetic and real datasets, and show that the proposed method outperforms other methods on most tasks, such as a nonlinear dataset with quadratic functions and a vector-valued dataset, while being competitive with the best method on other datasets.
\section{Background and Related Work}\label{pre}

\subsection{Structural Equation Model and Identifiability}\label{pre_scm}

Let $\mathcal{G}$ be a DAG with vertex set $V = \{X_1, X_2, \dots, X_d\}$ where each node $X_i$ represents a random variable. We denote by $X_{\mathrm{pa}(i)}$ the set of parental nodes of $X_i$ so that there is an edge from $X_j\in X_{\mathrm{pa}(i)}$ to $X_i$ in $\mathcal G$. In this work, we focus on the recursive SEM:
\begin{align}\label{scm}
X_i=f_i(X_{\mathrm{pa}(i)}) + \epsilon_i,\quad i=1,2,\dots, d, 
\end{align}
where $f_i$ is a deterministic function and $\epsilon_i$'s are jointly independent  noise variables with strictly positive densities (w.r.t.~Lebesgue measure). Such a form of SEM is called Additive Noise Model (ANM) and we use the two names interchangeably. We mostly consider nonlinear continuous functions for $f_i$'s and assume {causal minimality} which in this case reduces to that each $f_i$ is not a constant function in any input $X_j \in X_{\mathrm{pa}(i)}$ \citep{Peters2014causal}.  

Let $X$ be the vector concatenating all the variables $X_i$ and $P(X)$ the marginal distribution induced by the SEM defined on DAG $\mathcal G$. Then $P(X)$ is Markov w.r.t. $\mathcal{G}$, and $\mathcal{G}$ and  $P(X)$ are said to form a causal Bayesian network. The problem of causal structure learning is to use the observational data $\{x^{(k)}\}_{k=1}^n$, with $x^{(k)}$ being the $k$-th independent sample from the distribution $P(X)$, to infer the  causal graph $\mathcal G$. 

In general, however, it is impossible to recover $\cal G$ using only observational data from $P(X)$, because the underlying DAG is generally not identifiable without further assumption on the SEM \citep{Peters2014causal,Zhang2015estimation}. Hence, causal structure learning methods generally estimate, or whose output has to be converted to, the Markov equivalence class in order to have causal interpretation.  This fact relates to the \emph{identifiability} issue in causal structure learning: given an SEM defined on a DAG $\cal G$ with distribution $P(X)$, we say that $\mathcal G$ is \emph{identifiable} if no other SEMs can induce the same distribution $P(X)$ with a different DAG. Fortunately, \citet{Peters2014causal} have shown that if we consider only a subclass of SEMs, the {restricted ANMs} where $f_i$'s and the density functions of $\epsilon_i$'s and $X$ do not solve a system of three-order differential equations (see~\citet[Condition~19]{Peters2014causal}), then the true DAG is identifiable. Other identifiable models include linear non-Gaussian model \cite{Shimizu2006lingam}, linear Gaussian model with equal noise variances \cite{Peters2013identifiability}, post-nonlinear model \cite{Zhang2009identifiability}, etc. In this work, we will focus on the restricted ANMs for identifiability issues.

\subsection{Gradient-Based Structure Learning Methods}\label{sec:relatedwork}

NOTEARS \citep{zheng2018dags} formulates score-based structure learning of linear SEMs as a continuous optimization problem, based on a smooth characterization of acyclicity w.r.t.~the weighted adjacency matrix. With least squares loss, NOTEARS applies numeric methods to estimating the weighted adjacency matrix, followed by a thresholding step. DAG-GNN \cite{Yu19DAGGNN} and GAE \citep{Ng2019GAE} extend NOTEARS to nonlinear cases where a common nonlinear transformation is assumed to hold for each variable $X_i$, i.e., the nonlinear relationship applies to $X$ in a variable-wise manner. This assumption on the relationships may be restrictive in practice. GraN-DAG \cite{Lachapelle2019grandag} models conditional distribution of each variable given its parents with  Multi-Layer Perceptrons (MLPs) and constructs an equivalent weighted adjacency matrix based on MLP paths. A similar approach, NOTEARS-MLP \cite{Zheng2020learning}, uses partial derivatives to construct an equivalent weighted matrix; it involves choosing a suitable family of model functions and generally requires much effort in finding the parametric functions. Other related methods include \citet{Goudet2018learning} that uses generative neural networks for functional causal modeling with an initial graph skeleton, \citet{Ke2019learning} that uses logistic sigmoid functions to sample binary adjacency matrices in the presence of unknown interventions, \citet{Ng2020role} that adopts an unconstrained continuous formulation in the linear case with soft $\ell_1$ and DAG constraints, and \citet{Zhu2020causal,ijcai2021-491} that utilize policy gradient to find optimal DAGs without the requirement of  smooth score functions but may take a longer running time.

A more related work is SAM \cite{Klainathan2018sam} that also considered binary structural gates and the smooth acyclicity constraint. While certain similarities exist, we believe that there are more fundamental differences: (1) SAM studied functional causal models and discussed identification of Markov equivalence class. We formulate SEMs in a form parameterized by binary matrix and characterize the identification of super-graphs using the identifiability results of ANMs. (2) SAM considered a specific score function and used MLPs to fit causal relationships, which were trained in an adversarial way. We do not limit the form of score  and model functions, leading to a more general framework and training procedure. (3) running SAM, we find that its output matrix has many entries not near the boundary and may not be a DAG even after thresholding. We devise carefully the training procedure and stopping criterion for our method, resulting in an adjacency matrix  guaranteed to be acyclic after thesholding. As shown in Section~\ref{exp_gp}, these strategies make our method outperform SAM by a large margin.
\section{Structural Equation Model with Binary Adjacency Matrix}

Both NOTEARS and DAG-GNN rely on a notion of weighted adjacency matrix which does not exist for many SEMs, e.g., when the causal relationships are quadratic functions or functions sampled from Gaussian processes. Noticing that every directed graph corresponds to a binary adjacency matrix and vice versa, we consider another form of SEM that is explicitly parameterized by a binary adjacency matrix.

Let $A=[A_1|A_2|\cdots|A_d]$ be the binary adjacency matrix associated with the true DAG $\mathcal G$, where $A_i\in \{0, 1\}^{d}$ can be viewed as an indicator vector so that $A_{ji}$, the $j$-th entry of $A_i$, equals $1$ if and only if $X_j$ is a parent of $X_i$. To incorporate $A$ with Eq.~(\ref{scm}), we define a new function $g_i:\mathbb R^{d}\to\mathbb R$ so that the function value $g_i(x)=f_i(x_{\mathrm{pa}(i)})$ for each input $x$, i.e., $g_i(x)$ depends only on $x_{\mathrm{pa}(i)}$. We can then rewrite Eq.~(\ref{scm}) in a form parameterized by $A$:
\begin{align} \label{scm2}
X_i=g_i(A_i\circ X) + \epsilon_i, \quad i=1,2,\ldots, d,
\end{align}
where $\circ$ is the element-wise product. In this formulation, if $A_{ji}=0$, then  $X_i$ does not depend on the input $X_j$ anymore for any function $g_i$. Thus, $A_i$ can be viewed as a \emph{mask} to remove non-causal inputs.

Similar to existing gradient-based methods, it appears that we could use a parametric or nonparametric model, together with a $d\times d$ binary matrix, to
fit the observed data under the acyclicity constraint. With the fitted model $\hat g_i$ and $\hat A_i$, we may output the graph
indicated by $[\hat A_1|\hat A_2|\cdots|\hat A_d]$ as our estimated structure. Unfortunately, even if we can fit the data perfectly, it is
not clear whether the estimated binary matrix indicates the true causal graph: there may exist $g_i$, $A_i$ and $g_i'$, $A_i'$ that
induce the same distribution $P(X)$, but $A_i$ and $A_i'$ disagree for some $i$. In the following, we  show that such a formulation results in identification of a super-graph of the true graph under suitable conditions.

Formally, consider a marginal distribution $P(X)$ induced by a recursive SEM defined in Eq.~(\ref{scm}) with DAG $\mathcal G$. Assume that an SEM $X_i=h_i(B_i\circ X) + \tilde{\epsilon}_i,i=1,2,\ldots,d$ in the form of Eq.~(\ref{scm2}) induces the same marginal distribution $P(X)$, where $B_i\in\{0,1\}^d$, $B=[B_0|B_1|\cdots|B_d]$ represents a DAG $\cal H$, $h_i:\mathbb R^d\to \mathbb R$ is a function with input $X$, and $\tilde{\epsilon}_i$'s are jointly independent noises with strictly positive densities. We would like to ask: \emph{(1)~can $\mathcal H$ be uniquely identified from $P(X)$?} and \emph{(2)~what is the relationship of $\cal H$ to the true DAG~$\cal G$?}

Our answer is negative to the first question, because function $h_i$ can be degenerate w.r.t.~some of its input arguments. That is, if $h_i$ is a constant function w.r.t.~$X_j$, then whether $B_{ji}=0$ or $1$ does not affect $X_i$ but will change the structure of $\mathcal H$. We may then restrict $h_i$ to be non-degenerate w.r.t.~all $X_j,j\neq i$ to meet the causal minimality condition, but this restriction may be hard to place during learning. To proceed, let $X_{\widetilde{\mathrm{pa}}(i)}$ be the set of variables $X_j$ w.r.t.~which $h_i(B_i\circ X)$ is non-degenerate. We next define $\tilde f_i:  \mathbb R^{|X_{\widetilde{\mathrm{pa}}(i)}|} \to \mathbb R$ that maps $\tilde x\in\mathbb R^{|X_{\widetilde{\mathrm{pa}}(i)}|}$ to the function value $h_i(B_i\circ x)$ with $x$ being such that $x_{\widetilde{\mathrm{pa}}(i)}=\tilde x$. Such a mapping is possible because $h_i(B_i\circ x)$ depends only on $x_{\widetilde{\mathrm{pa}}(i)}$. It can be verified that the functions $\tilde f_i$'s and noises $\tilde\epsilon_i$'s then form an SEM satisfying causal minimality, and that the distribution of $X$ induced by this reduced SEM is identical to $P(X)$. Denoting by $\widetilde{\cal G}$ the causal DAG of the reduced SEM with functions $\tilde{f}_i$'s and noise variables $\tilde{\epsilon}_i$'s, we obtain Lemma \ref{lemma1} and its proof is given in Appendix~\ref{prooflemma1}.
\begin{lemma}\label{lemma1}
$\mathcal H$ is a super-graph of $\widetilde{\cal G}$, i.e., all the edges in $\widetilde{\cal G}$ also exist in $\mathcal H$. 
\end{lemma}

We proceed to the second question. Recall that the observed data are generated from a distribution induced by the SEM in Eq.~(\ref{scm}). We may further assume a restricted ANM  (see~\citet[Definition~27]{Peters2014causal}) for the data generating procedure so that the true causal graph $\cal G$ is identifiable. We then obtain the following proposition, with a proof given in Appendix~\ref{proofproposition1}.
\begin{proposition}\label{prop:identification}
Assume a restricted ANM with graph $\cal G$ and distribution $P(X)$ so that the original SEM is identifiable. If the parameterized SEM in the form of Eq.~(\ref{scm2}) with graph $\mathcal H$ induces the same $P(X)$, then $\cal H$ is a super-graph of $\cal G$.
\end{proposition}

With Proposition~\ref{prop:identification}, we may then apply parametric or nonparametric model functions and a binary adjacency matrix to fitting the SEM in the form of Eq.~(\ref{scm2}), subject to the acyclicity constraint. As in \citet[Proposition~1]{Lachapelle2019grandag}, if the causal relationships fall into the chosen model functions and we can obtain the exact solution that minimizes the negative log-likelihood given infinite samples, the resulting SEM has the same distribution. Consequently, we obtain an acyclic super-graph, from which existing nonlinear variable selection methods can be used to learn the parental sets and hence the causal graph. Such an idea has been exploited in many ordering based methods where fully-connected DAGs are first learned (see, e.g., \citet{teyssier2012ordering,Peters2014causal}). In practice, we can only apply approximate model functions and also need an efficient optimization procedure that can handle the binary adjacency matrix, which is the topic of next section.

\section{Masked Causal Structure Learning}
\label{sec:mask}

While we can adopt smooth model and score functions, the binary entries prohibit gradient-based optimization. A direct approach is to apply logistic sigmoid functions parameterized by real valued variables, which however lead to estimated entries lying in a small range near zero, making it hard to apply thresholding to identify edges (see Appendix~\ref{sec:sigmoidvsgumbelsigmoid} for an illustration). We would like each estimated entry to be either close to zero so that it almost removes a non-causal input, or near one so that we can easily apply a thresholding at $0.5$.

\subsection{Gumbel-Sigmoid for Binary Entries}
\label{mask_model}
We leverage the Gumbel-Softmax approach often used to approximate samples from a categorical distribution \citep{jang2017categorical,maddison2016concrete}. This approach has lower variances of gradient estimates than the straight-through method \citep{Bengio2013estimating} and also outperforms  several REINFORCE based methods \cite{Williams2004SimpleSG} on some
structured output prediction tasks. For a random variable  defined on $\{0,1\}$ with class probabilities $\pi_0\in(0,1)$ and $\pi_1=1-\pi_0$, its \emph{binary} sample can be approximated by
\begin{align}
\label{eqn:gumbel_sigmoid}
y & = \big(1 + \exp\,(-(\log(\pi_1/\pi_0)+ (\tilde{g}_1-\tilde{g}_0))/\tau)\big)^{-1}\nonumber\\
  & = \sigma((u + \tilde{g}) /\tau),
\end{align}
where $\tau>0$ is the temperature,  $\tilde{g}_1,\tilde{g}_2$ are independent samples from $\operatorname{Gumbel}(0,1)$, $\sigma(\cdot)$ is the logistic sigmoid function, and we define  $\tilde{g}=\tilde{g}_1-\tilde{g}_0$ and $u=\log(\pi_1/\pi_0)\in\mathbb R$. We refer to Eq.~(\ref{eqn:gumbel_sigmoid}) as Gumbel-Sigmoid with a logit $u$ and temperature $\tau$, and write it as $\mathsf{g}_\tau(u)$.

\begin{figure}
\vspace{-0.9em}
\centering
\subfloat[{$\tau=0.2$}]{
  \includegraphics[width=0.46\linewidth]{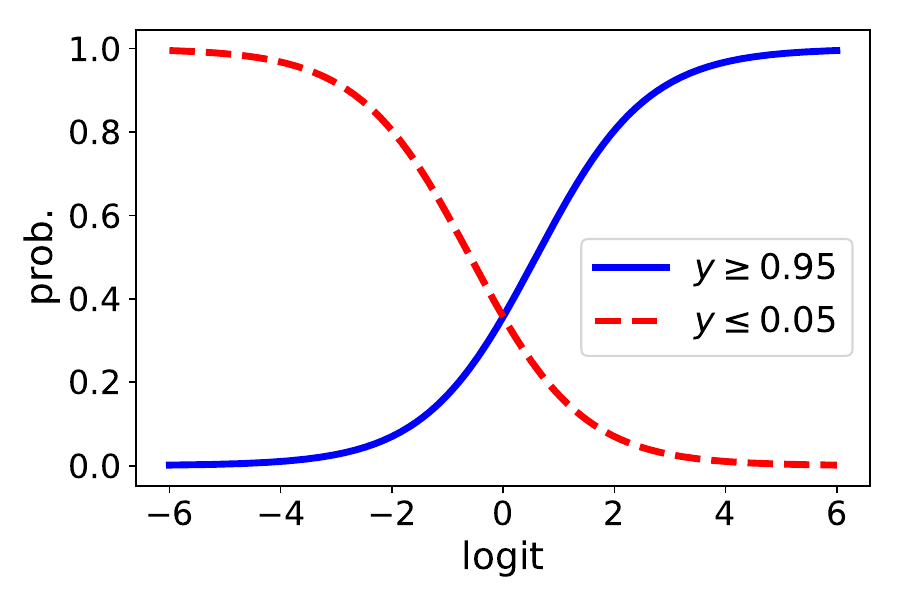}
  \label{fig:gum_sig1}
}
\subfloat[{$\operatorname{Pr}(y\leq0.05~\text{or}~y\geq0.95)$}]{
  \includegraphics[width=0.46\linewidth]{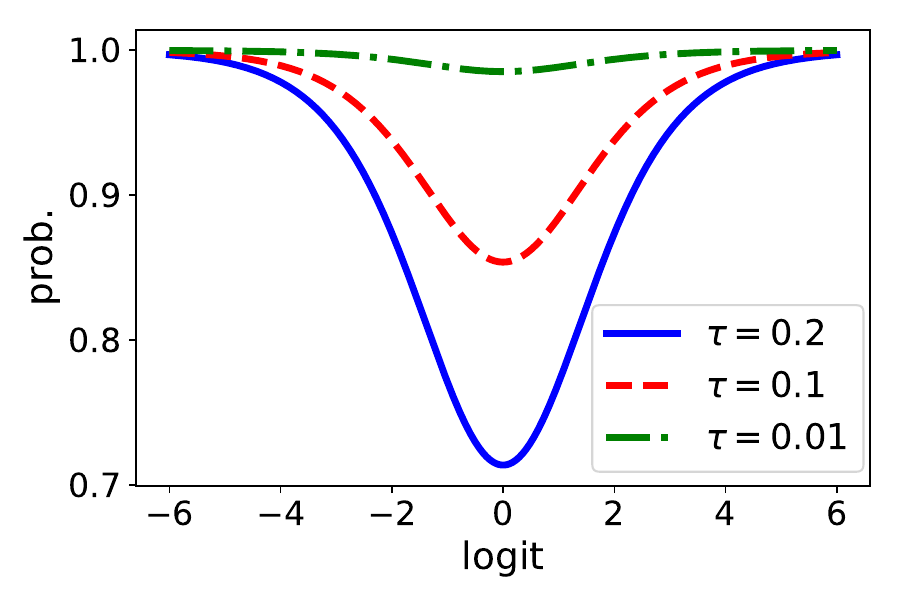}
  \label{fig:gum_sig2}
}
\caption{Gumbel-Sigmoid with different logits.} 
\label{fig:gubm_sig}
\vspace{-0.7em}
\end{figure}

It can also be shown that $\tilde{g}\sim\operatorname{Logistic}(0,1)$ (see Appendix~\ref{sec:gumbel_logistic} for a derivation). Figure~\ref{fig:gum_sig1} plots the probabilities  $\operatorname{Pr}(y\leq\delta)$ and $\operatorname{Pr}(y\geq1-\delta)$  for $\tau=0.2$ and $\delta=0.05$. The sum $\operatorname{Pr}\,(y\leq\delta)+ \operatorname{Pr}\,(y\geq1-\delta)$ characterizes the probability of a Gumbel-Sigmoid output lying within a neighborhood of $0$ or $1$, which goes to $1$ as $\tau\to0^+$ for any fixed $u$ and $\delta\in(0,0.5)$; see  Figure~\ref{fig:gum_sig2} for an illustration with $\delta=0.05$. In other words, a Gumbel-Sigmoid output can be arbitrarily close to $0$ or $1$ with high probability for a sufficiently small temperature.  In our experiments, we find that a small fixed $\tau$ (e.g., $0.2$)  works well.

We thus use Gumbel-Sigmoid to approximate the binary adjacency matrix. Consider that an edge helps produce a better reconstruction and does not violate the acyclicity constraint. Intuitively, if we repeatedly apply Gumbel-Sigmoid and estimate the score function between observed and reconstructed samples, gradient-based optimization will push the logit so that the Gumbel-Sigmoid output is close to one in the expected sense. If the acyclicity constraint is violated, some entries will be pushed towards zero to meet the constraint. 

\subsection{Acyclicity Constraint and Optimization}
\label{mask_acyclicity}
We use a similar smooth constraint from \citep{zheng2018dags} to enforce acyclicity. For a real matrix $U\in\mathbb R^{d\times d}$, we use $\mathsf{g}_\tau(U)$ to denote the output of applying Gumbel-Sigmoid to each entry $U_{ji}$ independently. As introduced in Section~\ref{mask_model}, $\mathsf{g}_\tau(U)$ is treated as our approximation to the binary adjacency matrix, and we further set the $(i,i)$-th entry of $\mathsf{g}_\tau(U)$ to zero to avoid self-loops. The acyclicity constraint reads $\mathbb E[\operatorname{tr}\left(e^{\,\mathsf{g}_\tau(U)}\right) - d]= 0$, which holds if and only if $\mathbb E[\mathsf{g}_\tau(U)]$ corresponds to a DAG \cite{zheng2018dags}. Here the expectation is taken w.r.t.~$\operatorname{Logistic}(0,1)$~samples independently for each entry and $e^M$ denotes matrix exponential of a square matrix $M$. While $\mathsf{g}_{\tau}(U_{ji})$ cannot be exactly $0$ and the constraint cannot be made exactly $0$, either, it suffices to make $\mathbb E[\operatorname{tr}\left(e^{\,\mathsf{g}_\tau(U)}\right) - d]<\xi$  for some small tolerance~$\xi$, followed by a thresholding step.

We now present our score-based structure learning method using the parameterized SEM in Eq.~(\ref{scm2}). With a slight abuse of notation, denote by $h_i:\mathbb R^{d}\to\mathbb R$ the function used to model causal relationships, i.e., we use $h_i(\mathsf{g}_\tau(U_i)\circ X; \phi_i)$ to reconstruct the variable $X_i$ where $U_i$ is the $i$-th column of $U$ and $\phi_i$ is the parameter associated with $h_i$. We have set the $i$-th element of $\mathsf{g}_\tau(U_i)$ to be $0$, and thus the output of $h_i$ will not be affected by $X_i$. Let $\phi=\{\phi_i\}_{i=1}^d$ and  $h(\mathsf{g}_\tau(U), X; \phi)=\{h_i(\mathsf{g}_\tau(U_i)\circ X);\phi_i\}_{i=1}^d$. With a score function $\mathcal L(\cdot, \cdot)$ defined w.r.t.~the observed and reconstructed samples, the constrained optimization problem is as follows:
\begin{align*}
\min_{U,\,\phi}~~&\mathbb E\bigg[\frac{1}{2n}\sum_{k=1}^{n}\mathcal L\Big(x^{(k)},h\left(\mathsf{g}_\tau(U), x^{(k)}; \phi\right)\hspace{-3pt}\Big)\hspace{-2pt}+\hspace{-2pt}\lambda\|\mathsf{g}_\tau(U)\|_{1}\bigg]\nonumber\\
\operatorname{s.t.}~~&\mathbb E\left[s(U)\right] \leq \xi,
\end{align*}
where expectations are taken w.r.t.~$\operatorname{Logistic}(0,1)$ samples and  $s(U)\coloneqq\operatorname{tr}\left(e^{\,\mathsf{g}_\tau(U)}\right) - d\geq 0$. A sparsity-inducing term $\|\mathsf{g}_\tau(U)\|_{1}$ is also incorporated.

Following \citet{zheng2018dags}, the above constrained optimization problem can be solved using augmented Lagrangian method when the score function $\mathcal L(\cdot, \cdot)$ and model functions $h_i$'s are chosen properly. For example, we may use the least squares loss or negative log-likelihood as the score function. The choices of $h_i$ include polynomial functions and MLPs. Augmented Lagrangian method consists of optimizing a sequence of subproblems where the exact solutions  converge to a stationary point of the original constrained problem under some regularity conditions \cite{Bertsekas/99}. The augmented Lagrangian of the above constrained problem is
\begin{align}
L_{\rho}(U,\phi, \alpha)= \mathbb E\bigg[&\frac{1}{2n}\sum_{k=1}^{n}\mathcal L\Big(x^{(k)},\, h\left(\mathsf{g}_\tau(U), x^{(k)}; \phi\right)\hspace{-3pt}\Big) \nonumber \\
&+\lambda\|\mathsf{g}_\tau(U)\|_{1}+\alpha s(U)\bigg]+\frac{\rho}{2}\big(\mathbb E[s(U)]\big)^{2},\nonumber
\end{align}
where $\rho>0$ is the penalty parameter and $\alpha$ is the estimate of Lagrange multiplier. For later use, we define $\mathbf g\in\mathbb R^{d\times d}$ as independent $\operatorname{Logistic}(0,1)$ samples associated with $U$ in $\mathsf{g}_\tau(U)$, and write $L_{\rho}(U,\phi, \alpha;\mathbf{g})$ and $s(U;\mathbf g)$ as sample estimates of $L_{\rho}(U,\phi, \alpha)$ and $\mathbb E\left[s(U)\right]$ evaluated with sample $\mathbf g$, respectively.

The updating rules for augmented Lagrangian method yield:
\begin{align}
\hspace{-0.05em}
  U^{t+1},\,\phi^{t+1} &= \argmin_{U,\,\phi} L_{\rho^{t}}(U, \phi, \alpha^t) \label{eqn:al_2}, \\
  \alpha^{t+1} &= \alpha^{t}+ \rho^{t}\,\mathbb E[s(U^{t+1})] \label{eqn:al_3}, \\
    \rho^{t+1} &=\begin{cases}
    \label{eqn:al_4}
    \beta \rho^{t}, &~\hspace{-0.5em}\text{if}~\,\mathbb E[s(U^{t+1})] \geq \gamma\mathbb E[s(U^{t})],\\  
    \rho^{t}, &~\hspace{-0.5em}\text{otherwise},
    \end{cases}
\end{align}
with $\beta > 1 $ and $0<\gamma < 1$ being the hyperparameters. Different from the optimization procedure in NOTEARS, the updating rules of Eqs.~(\ref{eqn:al_2}), (\ref{eqn:al_3}), and (\ref{eqn:al_4}) involve additional expectations w.r.t.~$\operatorname{Logistic}(0,1)$ distribution and their closed forms are not easy to obtain. Here we propose to use sample estimates as approximations and details are described below:
\begin{itemize}
\item The subproblem in Eq.~(\ref{eqn:al_2}) can be approximately solved using a gradient-based optimization algorithm, e.g., Adam \citep{Kingma2014adam}, where at each iteration we draw independently $\mathbf g_1, \mathbf g_2, \ldots, \mathbf g_b$ and  approximate the gradient by $\nabla_{U,\phi}\left(\frac1b\sum_{i=1}^b L_{\rho}(U,\phi, \alpha;\mathbf{g}_i)\right)$. 
\item With an approximate solution $U^{t+1}$ obtained from Eq.~(\ref{eqn:al_2}), we similarly use sample estimate to approximate $\mathbb E[s(U^{t+1})]$, which is  $\frac1b\sum_{i=1}^b s(U^{t+1};\mathbf{g}_i)$, in both Eqs.~(\ref{eqn:al_3}) and (\ref{eqn:al_4}).
\end{itemize}
As shown by \citet{jang2017categorical}, the Gumbel-Softmax approach is effective for single sample gradient estimation, so we simply pick $b=1$ which is found to work well.

Augmented Lagrangian method usually stops once the constraint is satisfied. In our case, while a single sample estimation of gradient works well for training, it may render the optimization to stop too early if a sample `accidentally' makes the constraint below the tolerance. Thus, we choose the tolerance $\xi$ to be small and additionally use $\operatorname{tr}(e^{\sigma(U/\tau)})-d < \xi$  as our stopping criterion to lower the probability of stopping the algorithm early.  Moreover, notice that the latter criterion is satisfied only when the entries $\sigma(U_{ji}/\tau)$, where $i,j$ are such that the edge $X_j\to X_i\notin E$ for some DAG with $E$ as edge set, are nearly zeros. If an edge indeed helps minimize the score function,  the logit will be pushed to a relatively large positive value so that the score function is minimized in the expected sense. For edges that do not violate acyclicity nor help minimize the score function, the corresponding entries may have intermediate values. Nevertheless, these edges are treated as spurious edges and will be further processed. Thus, as a byproduct of introducing a second stopping criterion, we can readily use $\sigma(U/\tau)$ as our learned matrix, which would indicate a DAG after thresholding at  $0.5$.

More details regarding the parameter choices, stopping criterion, and overfitting issue can be found in Appendix~\ref{sec:optimization}. We remark that such a  stopping criterion, together with the devised optimization procedure, leads to a much better performance than the related method SAM \cite{Klainathan2018sam}; see experimental results in Section~\ref{exp_gp}.

\subsection{Final Output}
A learned matrix $\sigma(U/\tau)$ after thresholding may contain spurious edges or false discoveries. The $\ell_1$ penalty term can be used to reduce false discoveries, yet picking a proper penalty weight is not easy. We therefore stick to a small weight (i.e., $2\times 10^{-3}$) to slightly control false discoveries during training, followed by variable selection to obtain the final estimate. A practically useful  approach is the Causal Additive Model (CAM) based pruning method proposed by \citet{Buhlmann2014cam}, whose performance has also been validated by \citet{Lachapelle2019grandag,Zheng2020learning}. Appendix~\ref{exp_pruning} provides more details regarding CAM pruning and its effect on our method.
\begin{table*}[t]
\centering
\caption{Empirical results on nonlinear SEMs with Gaussian processes. Lower SHD and higher TPR are better.}
\vspace{-0.5em}
\label{tab:results_gp}{\notsotiny
\begin{tabular}{lll|ll|ll|llllll}
\toprule \vspace{-0.1em}
~& \multicolumn{2}{l}{ER1 with $10$ nodes} & \multicolumn{2}{l}{ER4 with $10$ nodes} 
    & \multicolumn{2}{l}{ER1 with $50$ nodes} & \multicolumn{2}{l}{ER4 with $50$ nodes} \\
    \cmidrule(lr){2-3} \cmidrule(lr){4-5} \cmidrule(lr){6-7} \cmidrule(lr){8-9}
& \multicolumn{1}{l}{SHD} & \multicolumn{1}{l}{TPR} & \multicolumn{1}{l}{SHD} & \multicolumn{1}{l}{TPR}
    & \multicolumn{1}{l}{SHD} & \multicolumn{1}{l}{TPR} & \multicolumn{1}{l}{SHD} & \multicolumn{1}{l}{TPR}
\\ \midrule
MCSL-MLP   &    {\bf 1.4\,$\pm$\,1.7}  & {\bf 0.87\,$\pm$\,0.21} &  {\bf 8.2\,$\pm$\,3.3}   &  {\bf 0.80\,$\pm$\,0.09} 
    &   9.2\,$\pm$\,4.8 & {\bf 0.89\,$\pm$\,0.05} &  {\bf 49.6\,$\pm$\,12.2} & {\bf 0.78\,$\pm$\,0.04} \\
GraN-DAG   &    {\bf 2.4\,$\pm$\,2.2}  & {0.86\,$\pm$\,0.15}  & {\bf 14.4\,$\pm$\,4.8}   &  {\bf 0.66\,$\pm$\,0.12}
    &   {\bf 6.6\,$\pm$\,3.5} & {\bf 0.92\,$\pm$\,0.05} &  {\bf 59.4\,$\pm$\,12.9}& {\bf 0.75\,$\pm$\,0.05} \\
CAM        &    5.0\,$\pm$\,2.3  & {\bf 0.92\,$\pm$\,0.08}   & {\bf 16.6\,$\pm$\,3.4}   &  \bf{0.63\,$\pm$\,0.08}
    &  {\bf 3.8\,$\pm$\,1.9} & {\bf 0.96\,$\pm$\,0.02} &  {\bf 58.6\,$\pm$\,6.6} & {\bf 0.76\,$\pm$\,0.02} \\
NOTEARS-MLP~&    {\bf 1.8\,$\pm$\,1.3} & {\bf 0.87\,$\pm$\,0.12} & 25.4\,$\pm$\,4.7  & 0.40\,$\pm$\,0.10
    &  {\bf 6.8\,$\pm$\,1.9} & {\bf 0.89\,$\pm$\,0.03} & 119.0\,$\pm$\,17.0 & 0.43\,$\pm$\,0.06\\
SAM     &    6.4\,$\pm$\,1.6  & 0.67\,$\pm$\,0.10   & 33.8\,$\pm$\,3.6  &  0.20\,$\pm$\,0.09
    &  35.8\,$\pm$\,7,4 & 0.46\,$\pm$\,0.08 & 192.2\,$\pm$\,10.0 & 0.07\,$\pm$\,0.02\\
DAG-GNN    &    6.6\,$\pm$\,3.6  & 0.50\,$\pm$\,0.24  & 37.0\,$\pm$\,2.1   &  0.12\,$\pm$\,0.03
    &  32.2\,$\pm$\,7.8 & 0.45\,$\pm$\,0.10 & 186.2\,$\pm$\,14.5& 0.10\,$\pm$\,0.04 \\
NOTEARS    &    5.0\,$\pm$\,2.9  & 0.62\,$\pm$\,0.19   & 35.2\,$\pm$\,2.5   &  0.15\,$\pm$\,0.05
    &  22.8\,$\pm$\,7.0 & 0.67\,$\pm$\,0.10 & 174.8\,$\pm$\,13.5& 0.16\,$\pm$\,0.03 \\
GES        &    3.4\ $\pm$\,1.7 & 0.78\,$\pm$\,0.13   & 29.4\,$\pm$\,1.0   &  0.30\,$\pm$\,0.30
    &  19.0\,$\pm$\,6.6 & 0.74\,$\pm$\,0.09 & 147.4\,$\pm$\,16.0& 0.31\,$\pm$\,0.05\\
PC         &    4.8\,$\pm$\,1.9  & 0.69\,$\pm$\,0.09   & 18.8\,$\pm$\,3.54  &  0.55\,$\pm$\,0.10
    &  28.0\,$\pm$\,6.5 & 0.67\,$\pm$\,0.07 & 123.2\,$\pm$\,11.4& 0.43\,$\pm$\,0.03\\
GES-GS     &    4.4\ $\pm$\,4.5 & 0.74\,$\pm$\,0.24   &  25.0\,$\pm$\,3.4   &  0.41\,$\pm$\,0.07
    &  20.4\,$\pm$\,4.0 & 0.76\,$\pm$\,0.04 & 154.0\,$\pm$\,18.9 & 0.32\,$\pm$\,0.06\\
PC-KCI     &    6.7\,$\pm$\,3.1  & 0.83\,$\pm$\,0.31   & 29.8\,$\pm$\,2.61  &  0.29\,$\pm$\,0.06
    &  41.2\,$\pm$\,5.58 & 0.61\,$\pm$\,0.13 & N/A & N/A     \\
\bottomrule
\end{tabular}}
\vspace{-0.5em}
\end{table*}

\section{Experiments}\label{exp}
We compare the proposed method against several baselines, including PC \citep{spirtes1991algorithm}, PC-KCI (with kernel-based conditional independence tests) \citep{Zhang2012kernel}, GES \citep{chickeringo2002optimal},  GES-GS (with generalized score) \citep{Huang2018generalized}, CAM \citep{Buhlmann2014cam}, NOTEARS \citep{zheng2018dags}, DAG-GNN \citep{Yu19DAGGNN}, SAM \cite{Klainathan2018sam}, NOTEARS-MLP \citep{Zheng2020learning}, and GraN-DAG \citep{Lachapelle2019grandag}. Here, our method uses $4$-layer MLPs as the model functions and is denoted as MCSL-MLP. Detailed setting of the MLP and hyperparameters is described in Appendix~\ref{sec:hyperparameters}. We use least squares as our score function in all experiments. An implementation of MCSL has been released in the \texttt{gCastle} package \cite{gcastle}.\footnote{\url{https://github.com/huawei-noah/trustworthyAI}}

For synthetic data, experiments are conducted on data models varying along graph size, degree, and causal relationship. We sample a DAG $\mathcal G$ using the Erd\"{o}s--R\'{e}nyi (ER) model and generate data in the causal order indicated by $\mathcal G$. We consider $d$-node ER graphs with on average $d$ and $4d$ edges, denoted as ER1 and ER4, respectively. Unless otherwise stated, we set the sample size to $3000$ and the additive noises to standard Gaussians. The causal relationships include functions sampled from Gaussian processes (Section~\ref{exp_gp}),  quadratic functions (Section~\ref{exp_quadratic}), and post-nonlinear models (Section~\ref{sec:exp_pnl}). A vector-valued case is considered in Section~\ref{vector}. For real data, we conduct experiments on a protein and a telecommunication dataset in Section \ref{sec:results_sachs}.

We further conduct empirical studies with scale-free graphs and different sample sizes; the detailed results are provided in Appendix~\ref{sec:scale_free_exp} and \ref{sec:consistency}, respectively. A discussion of the training time and computational complexity is available in Appendix~\ref{sec:complexity}.

We report Structural Hamming Distance (SHD) and True Positive Rate (TPR) to evaluate the learned graphs, averaged over five random seeds. For GES (GES-GS) and PC (PC-KCI), we treat undireted edges as true positives if the true graph has a directed edge in place of the undirected ones.

\begin{figure*}
	\minipage{0.32\textwidth}
	\includegraphics[width=\linewidth]{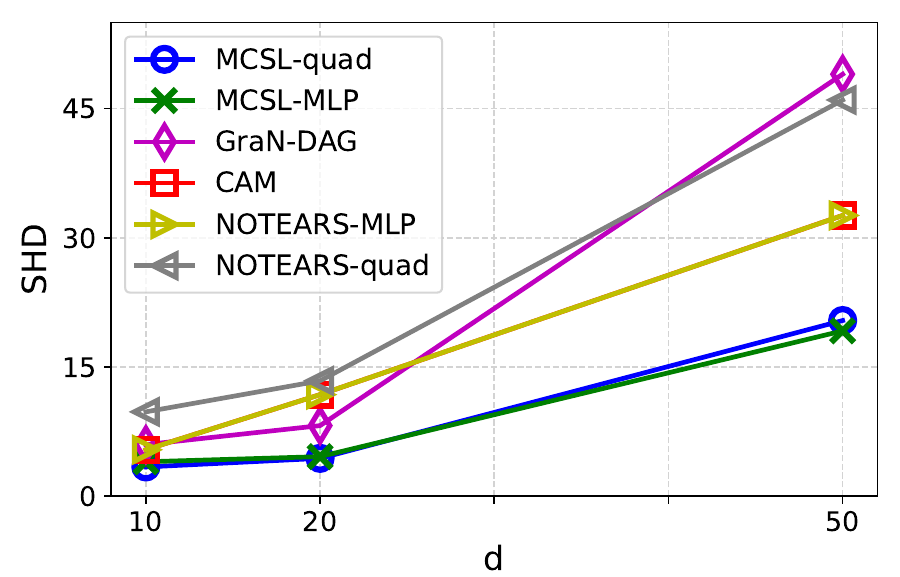}
	\caption{Quadratic functions.}\label{fig:results_quadratic}
	\endminipage\hfill
	\minipage{0.32\textwidth}
	\includegraphics[width=\linewidth]{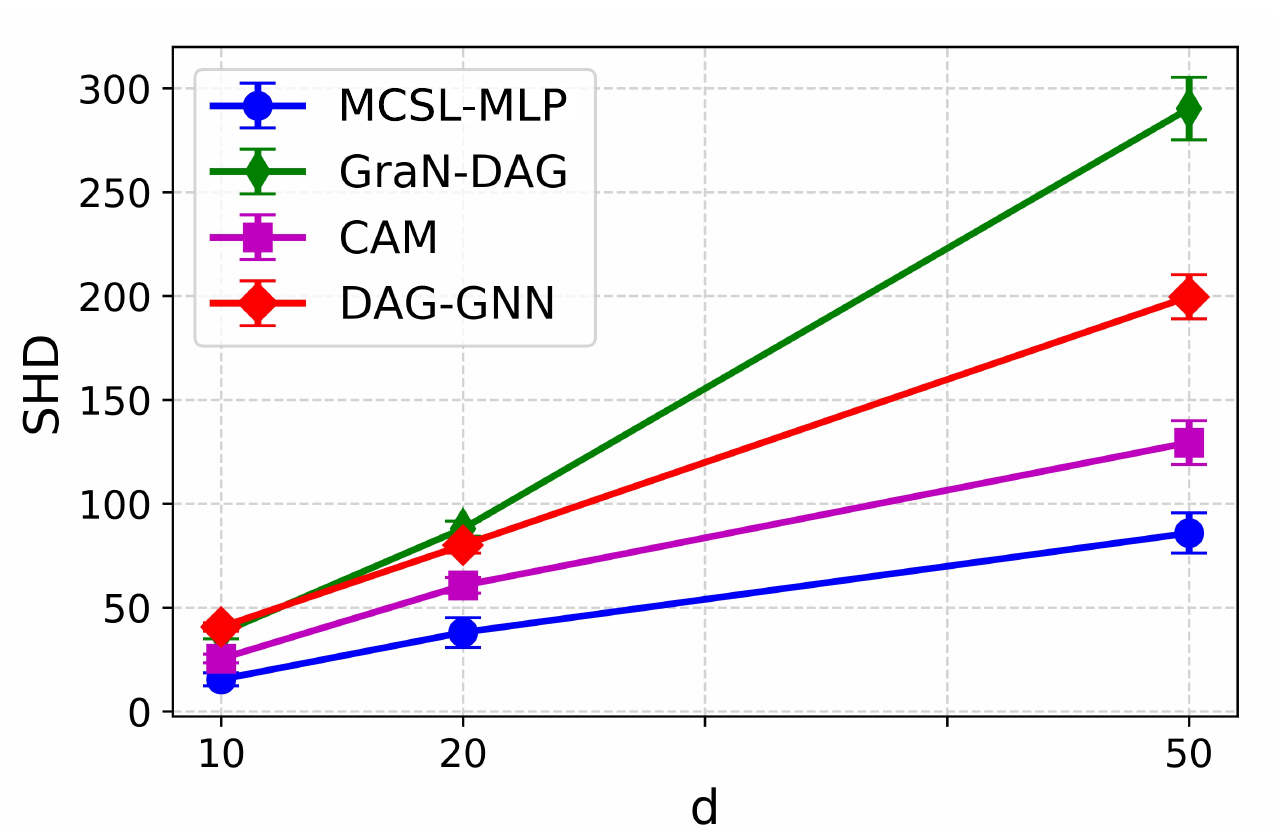}
	\caption{Vector-valued case.}\label{fig:vector}
	\endminipage\hfill
	\minipage{0.32\textwidth}
	\includegraphics[width=\linewidth]{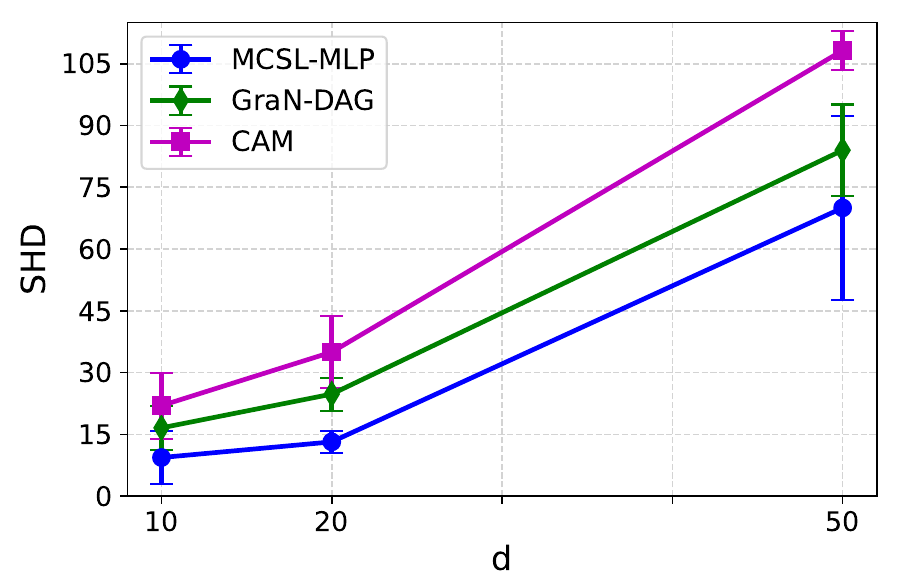}
	\caption{Post-nonlinear models.}\label{fig:results_pnl}
	\endminipage
	\vspace{-0.9em}
\end{figure*}

\subsection{Gaussian Processes} \label{exp_gp}
We first consider the data model previously used by \citet{Peters2014causal,Lachapelle2019grandag,Zheng2020learning}: each function $f_i$ is sampled from a Gaussian Process (GP) with RBF kernel of bandwidth one.  This setting is known to be  identifiable  \cite{Peters2014causal}. For MCSL-MLP, GraN-DAG, CAM, and NOTEARS-MLP, an additional CAM pruning step is used to remove spurious edges. We did not apply this pruning step to other baselines as they have much lower TPRs, especially on ER4 graphs.

The empirical results are reported in Table~\ref{tab:results_gp} with graph sizes $d\in\{10, 50\}$ and the results for $d\in\{20, 100\}$ are given in Appendix~\ref{more_exp_gp}. We observe that MCSL-MLP, GraN-DAG, and CAM outperform the other methods across most settings. MCSL-MLP has a better performance with small graphs while CAM performs the best for $100$-node graphs. Nevertheless, their differences are minor, compared with the performance of other methods. NOTEARS-MLP is on par with MCSL-MLP, GraN-DAG, and CAM on ER1 graphs, but performs poorly on ER4 graphs.  Both DAG-GNN and NOTEARS have poor performance, possibly because they cannot model this type of causal relationships, and moreover, they operate on the notion of weighted adjacency matrix which is not obvious here. Kernel-based methods have a cubic computational complexity w.r.t.~sample 
size, so we pick $500$ samples for each dataset. Yet they may still be too slow for large ER4 graphs ($>12$ hours) and we do not report their results for these cases. We observe that GES-GS and PC-KCI perform similarly to GES and PC, respectively. A possible reason is that although more powerful score function and conditional independence test are utilized, GES-GS and PC-KCI are limited by their computational complexity and only part of the samples can be used. Finally, we find that the output graph of SAM is not guaranteed to be acyclic and the performance is outperformed by a large margin; in particular, it learns very few true edges with $50$-node ER4 graphs.

\subsection{Quadratic Functions}\label{exp_quadratic}
We consider nonlinear relationships with quadratic functions, as in \citet{Zhu2020causal}. This data model is also  identifiable  \citep{Peters2014causal}.  With prior knowledge that the causal relationship follows a quadratic function, \citet{Zhu2020causal} used a similar idea from GraN-DAG to modify NOTEARS by constructing an equivalent weighted adjacency matrix. We refer to this method as NOTEARS-quad in this work and omit its details here; see \citet[Appendix~E]{Zhu2020causal} for a detailed description. A Quadratic Regression based pruning method (QR pruning) has also been shown to perform well for this data model, so we use it for all methods here. Details for experiment setup and QR pruning can be found in Appendix~\ref{more_exp_quadratic}. 

We may also utilize the prior knowledge to apply quadratic regressions to modeling causal relationships and denote the resulting method as MCSL-quad. For NOTEARS-quad, there is extra effort to derive explicitly the equivalent adjacency matrix and further re-implement the method. By contrast, our method only requires replacing the MLPs with quadratic functions.

Here we only consider methods that perform well in the experiment from Section \ref{exp_gp}. For better visualization, we report the average SHDs in Figure~\ref{fig:results_quadratic}, with detailed results including standard deviations and  TPRs given in Appendix~\ref{more_exp_quadratic}. MCSL-MLP and MCSL-quad have a similar performance and outperform the other methods by a large margin. This also shows that MCSL-MLP has a strong expressive power, as its performance is on par with MCSL-quad that utilizes the knowledge of the form of causal relationships. Although NOTEARS-quad is specifically designed for quadratic functions, it 
performs poorly here. We believe that it is because the equivalent adjacency matrix makes the optimization problem more complicated and difficult to solve.

\subsection{Vector-Valued Case}\label{vector}
We next demonstrate the utility of binary adjacency matrix to deal with vector-valued case, where it is straightforward to apply masking to each dimension of the variable.  With a little abuse of notation, we now let each $X_i$ be an $m$-dim  variable. To construct a vector-valued dataset, we extend the RBF kernel by defining $k(x,y) = \exp(-\|x-y\|_{\mathrm F}^2/2)$ for $x, y\in\mathbb R^{|X_{\mathrm{pa}(i)}|\times m}$, and then  sample a function from the GP with this kernel as causal relationship for each dimension of variable $X_i$. Here we pick $m=3$ and consider ER4 graphs.

For vector-valued case, we only need to modify the input and output dimensions of the MLPs in MCSL-MLP. CAM pruning only works with scalar variables, so we increase the $\ell_1$ penalty weight to $0.02$ to control false discoveries. GraN-DAG and CAM do not work for vector-valued case and here we pick the first dimension of each variable as observational data. Figure~\ref{fig:vector} shows the SHDs, where MCSL-MLP performs the best, thanks to its flexibility of including different model functions. Interestingly, CAM still has a relatively good performance. Although DAG-GNN is developed to include vector-valued case,  a notation of weighted adjacency matrix in the SEM leads to a poor performance.

\subsection{Post-Nonlinear Causal Models} 
\label{sec:exp_pnl}
To examine the performance of our method when the restricted ANM assumption is not met, we consider a post-nonlinear causal model given by $X_i= \exp\big(\log\sum_{X_j \in X_{\mathrm{pa}(i)}} X_j - \log|X_{\mathrm{pa}(i)}|+\epsilon_i\big)$, where noise variable $\epsilon_i$ follows the half-Gaussian distribution with unit variance. This setting is adapted from \citet{Zhang2015estimation}; \citet{Lachapelle2019grandag} and known to be identifiable. Note that this data model does also not satisfy the ANM assumption required by GraN-DAG and CAM owing to the post-nonlinearity, thus giving rise to mild model misspecification. We consider ER1 graphs with $d\in\{10, 20, 50\}$ nodes. The average SHDs are depicted in Figure \ref{fig:results_pnl}, with complete results available in Appendix \ref{sec:more_exp_pnl}, showing that MCSL-MLP performs the best across different graph sizes, as comapred to GraN-DAG and CAM with very low TPRs. This demonstrates that the proposed method seems to generalize well to cases where the assumption of restricted ANM is not met.

\subsection{Real Data}\label{sec:results_sachs}
\paragraph{Sachs dataset} We consider to estimate a protein signaling network based on expression levels of proteins and phospholipid \citep{Sachs2005causal}, containing both interventional and observational data. Since our method is based on passively observed data, we use only the observational data with $853$ samples. The ground truth proposed by \citet{Sachs2005causal} has $11$ nodes and $17$ edges. MCSL-MLP and CAM achieve the best SHD $12$, and GraN-DAG has an SHD $13$. DAG-GNN and NOTEARS are on par with MCSL-MLP and CAM w.r.t.~true positives, but have more false discoveries, resulting in SHDs $16$ and $19$, respectively. GES obtains $7$ undirected edges while PC estimates $7$ undirected and $1$ directed edges. By contrast, the inferred graphs from MCSL-MLP, CAM, and GraN-DAG consist of only directed edges.

\paragraph{Identifying root cause in telecommunication networks}We proceed to apply the proposed approach to identifying root cause of anomaly variables with time series data. Here each variable represents a certain key performance indicator (like number of users per minute) monitoring the state of telecommunication system. In practice, it is commonly observed that a single system fault may first happen and incur a variable to behave abnormally, and then the variable would cause several others to be abnormal. Thus, we would like to find the root cause one among all abnormal variables, which is believed to be more informative to identify the system fault. This task can be done by learning a causal DAG and then choose the root node. If there are more than one root nodes, our strategy is to return the one with the most descendants.

For time series data, several methods \cite{runge2019detecting,peters2013causal,chu2008search} first choose a time lag $p\geq 0$ and  then perform causal discovery among $d(p+1)$ variables, where many edges can be excluded as the variable at the current time cannot be a cause of the one of past time. Thus, GraN-DAG, CAM, and NOTEARS-MLP do not easily apply here and we can only consider instantaneous data, i.e., treat time series data as i.i.d.~data. Besides, prior knowledge from human experience indicates that the summary graph is acyclic and also some variables cannot be the root cause ones. Most existing approaches, however,  do not take into account this condition and directly learn the graph among $d(p+1)$ variables, followed by compressing the graph into a summary graph of $d$ variables. 

For this task, we can easily adapt our method as in the vector-valued case. Notice that 
this problem may also contain hidden confounders and in fact is very challenging. Nonetheless, in all our test cases containing $2$ to $60$ variables, our modified method can identify about $80\%$ correct root causes. For the reasons mentioned above, all the other methods can only find correctly at most $30\%$ cases; in particular, constraint-based methods usually result in many undirected edges, making it hard to determine a root~node.
\section{Concluding Remarks}
In this work, we reformulate the SEM to incorporate binary adjacency matrix and investigate the structure identification issue. A gradient-based  method is proposed by leveraging the recently developed  Gumbel-Softmax approach and the smooth acyclicity constraint. Experiments validate the effectiveness as well as the flexibility to include different model functions to fit the underlying causal relationships. A future direction is to use preliminary variable selection and/or second-order optimization to improve both the efficiency and efficacy.

{\small 
\bibliographystyle{abbrvnat}
\bibliography{bib/bibliography.bib}}

\begin{thebibliography}{45}
\providecommand{\natexlab}[1]{#1}
\providecommand{\url}[1]{\texttt{#1}}
\expandafter\ifx\csname urlstyle\endcsname\relax
  \providecommand{\doi}[1]{doi: #1}\else
  \providecommand{\doi}{doi: \begingroup \urlstyle{rm}\Url}\fi

\bibitem[Bengio et~al.(2013)Bengio, L{\'e}onard, and
  Courville]{Bengio2013estimating}
Y.~Bengio, N.~L{\'e}onard, and A.~Courville.
\newblock Estimating or propagating gradients through stochastic neurons for
  conditional computation.
\newblock \emph{arXiv preprint arXiv:1308.3432}, 2013.

\bibitem[Bertsekas(1999)]{Bertsekas/99}
D.~P. Bertsekas.
\newblock \emph{Nonlinear Programming}.
\newblock Athena Scientific, 1999.

\bibitem[B{\"u}hlmann et~al.(2014)B{\"u}hlmann, Peters, Ernest,
  et~al.]{Buhlmann2014cam}
P.~B{\"u}hlmann, J.~Peters, J.~Ernest, et~al.
\newblock {CAM}: Causal additive models, high-dimensional order search and
  penalized regression.
\newblock \emph{The Annals of Statistics}, 42\penalty0 (6):\penalty0
  2526--2556, 2014.

\bibitem[Chickering(1996)]{Chickering1996learning}
D.~M. Chickering.
\newblock Learning {Bayesian} networks is {NP}-complete.
\newblock In \emph{Learning from Data: Artificial Intelligence and Statistics
  V}. Springer, 1996.

\bibitem[Chickering(2002)]{chickeringo2002optimal}
D.~M. Chickering.
\newblock Optimal structure identification with greedy search.
\newblock \emph{Journal of Machine Learning Research}, 3\penalty0
  (Nov):\penalty0 507--554, 2002.

\bibitem[Chu and Glymour(2008)]{chu2008search}
T.~Chu and C.~Glymour.
\newblock Search for additive nonlinear time series causal models.
\newblock \emph{Journal of Machine Learning Research}, 9\penalty0
  (May):\penalty0 967--991, 2008.

\bibitem[Glorot and Bengio(2010)]{Glorot2010TrainingDNN}
X.~Glorot and Y.~Bengio.
\newblock Understanding the difficulty of training deep feedforward neural
  networks.
\newblock In \emph{AISTATS}, 2010.

\bibitem[Glymour et~al.(2019)Glymour, Zhang, and Spirtes]{Glymour2019review}
C.~Glymour, K.~Zhang, and P.~Spirtes.
\newblock Review of causal discovery methods based on graphical models.
\newblock \emph{Frontiers in Genetics}, 10, 2019.

\bibitem[Goudet et~al.(2018)Goudet, Kalainathan, Caillou, Guyon, Lopez-Paz, and
  Sebag]{Goudet2018learning}
O.~Goudet, D.~Kalainathan, P.~Caillou, I.~Guyon, D.~Lopez-Paz, and M.~Sebag.
\newblock Learning functional causal models with generative neural networks.
\newblock In \emph{Explainable and Interpretable Models in Computer Vision and
  Machine Learning}. Springer, 2018.

\bibitem[Huang et~al.(2018)Huang, Zhang, Lin, Sch{\"o}lkopf, and
  Glymour]{Huang2018generalized}
B.~Huang, K.~Zhang, Y.~Lin, B.~Sch{\"o}lkopf, and C.~Glymour.
\newblock Generalized score functions for causal discovery.
\newblock In \emph{KDD}, 2018.

\bibitem[Jang et~al.(2017)Jang, Gu, and Poole]{jang2017categorical}
E.~Jang, S.~Gu, and B.~Poole.
\newblock Categorical reparameterization with {Gumbel-Softmax}.
\newblock In \emph{ICLR}, 2017.

\bibitem[Kalainathan et~al.(2018)Kalainathan, Goudet, Guyon, Lopez-Paz, and
  Sebag]{Klainathan2018sam}
D.~Kalainathan, O.~Goudet, I.~Guyon, D.~Lopez-Paz, and M.~Sebag.
\newblock Structural agnostic modeling: Adversarial learning of causal graphs.
\newblock \emph{arXiv preprint arXiv:1803.04929}, 2018.

\bibitem[Ke et~al.(2019)Ke, Bilaniuk, Goyal, Bauer, Larochelle, Pal, and
  Bengio]{Ke2019learning}
N.~R. Ke, O.~Bilaniuk, A.~Goyal, S.~Bauer, H.~Larochelle, C.~Pal, and
  Y.~Bengio.
\newblock Learning neural causal models from unknown interventions.
\newblock \emph{arXiv preprint arXiv:1910.01075}, 2019.

\bibitem[Kingma and Ba(2014)]{Kingma2014adam}
D.~P. Kingma and J.~Ba.
\newblock Adam: A method for stochastic optimization.
\newblock \emph{ICLR}, 2014.

\bibitem[Koivisto and Sood(2004)]{Koivisto2004exact}
M.~Koivisto and K.~Sood.
\newblock Exact {Bayesian} structure discovery in {Bayesian} networks.
\newblock \emph{Journal of Machine Learning Research}, 5\penalty0
  (May):\penalty0 549--573, 2004.

\bibitem[Koller and Friedman(2009)]{Koller2009}
D.~Koller and N.~Friedman.
\newblock \emph{Probabilistic Graphical Models: Principles and Techniques}.
\newblock MIT Press, 2009.

\bibitem[Lachapelle et~al.(2020)Lachapelle, Brouillard, Deleu, and
  Lacoste-Julien]{Lachapelle2019grandag}
S.~Lachapelle, P.~Brouillard, T.~Deleu, and S.~Lacoste-Julien.
\newblock Gradient-based neural {DAG} learning.
\newblock In \emph{ICLR}, 2020.

\bibitem[Maddison et~al.(2017)Maddison, Mnih, and Teh]{maddison2016concrete}
C.~J. Maddison, A.~Mnih, and Y.~W. Teh.
\newblock The concrete distribution: A continuous relaxation of discrete random
  variables.
\newblock In \emph{ICLR}, 2017.

\bibitem[Meek(1995)]{meek1995casual}
C.~Meek.
\newblock Causal inference and causal explanation with background knowledge.
\newblock In \emph{UAI}, 1995.

\bibitem[Ng et~al.(2019)Ng, Zhu, Chen, and Fang]{Ng2019GAE}
I.~Ng, S.~Zhu, Z.~Chen, and Z.~Fang.
\newblock A graph autoencoder approach to causal structure learning.
\newblock \emph{arXiv preprint arXiv:1911.07420}, 2019.

\bibitem[Ng et~al.(2020)Ng, Ghassami, and Zhang]{Ng2020role}
I.~Ng, A.~Ghassami, and K.~Zhang.
\newblock On the role of sparsity and {DAG} constraints for learning linear
  {DAGs}.
\newblock In \emph{NeurIPS}, 2020.

\bibitem[Pearl(2009)]{pearl2009causality}
J.~Pearl.
\newblock \emph{Causality}.
\newblock Cambridge University Press, 2009.

\bibitem[Peters and B{\"u}hlmann(2013)]{Peters2013identifiability}
J.~Peters and P.~B{\"u}hlmann.
\newblock Identifiability of {Gaussian} structural equation models with equal
  error variances.
\newblock \emph{Biometrika}, 101\penalty0 (1):\penalty0 219--228, 2013.

\bibitem[Peters et~al.(2013)Peters, Janzing, and
  Sch{\"o}lkopf]{peters2013causal}
J.~Peters, D.~Janzing, and B.~Sch{\"o}lkopf.
\newblock Causal inference on time series using restricted structural equation
  models.
\newblock In \emph{NeurIPS}, 2013.

\bibitem[Peters et~al.(2014)Peters, Mooij, Janzing, and
  Sch{\"o}lkopf]{Peters2014causal}
J.~Peters, J.~M. Mooij, D.~Janzing, and B.~Sch{\"o}lkopf.
\newblock Causal discovery with continuous additive noise models.
\newblock \emph{The Journal of Machine Learning Research}, 15\penalty0
  (1):\penalty0 2009--2053, 2014.

\bibitem[Peters et~al.(2017)Peters, Janzing, and Sch{\"o}lkopf]{Peters2017book}
J.~Peters, D.~Janzing, and B.~Sch{\"o}lkopf.
\newblock \emph{Elements of Causal Inference - Foundations and Learning
  Algorithms}.
\newblock MIT Press, Cambridge, MA, USA, 2017.

\bibitem[Runge et~al.(2019)Runge, Nowack, Kretschmer, Flaxman, and
  Sejdinovic]{runge2019detecting}
J.~Runge, P.~Nowack, M.~Kretschmer, S.~Flaxman, and D.~Sejdinovic.
\newblock Detecting and quantifying causal associations in large nonlinear time
  series datasets.
\newblock \emph{Science Advances}, 5\penalty0 (11), 2019.

\bibitem[Sachs et~al.(2005)Sachs, Perez, Pe'er, Lauffenburger, and
  Nolan]{Sachs2005causal}
K.~Sachs, O.~Perez, D.~Pe'er, D.~A. Lauffenburger, and G.~P. Nolan.
\newblock Causal protein-signaling networks derived from multiparameter
  single-cell data.
\newblock \emph{Science}, 308\penalty0 (5721):\penalty0 523--529, 2005.

\bibitem[Shimizu et~al.(2006)Shimizu, Hoyer, Hyv{\"a}rinen, and
  Kerminen]{Shimizu2006lingam}
S.~Shimizu, P.~O. Hoyer, A.~Hyv{\"a}rinen, and A.~Kerminen.
\newblock A linear {non-Gaussian} acyclic model for causal discovery.
\newblock \emph{Journal of Machine Learning Research}, 7\penalty0
  (Oct):\penalty0 2003--2030, 2006.

\bibitem[Singh and Moore(2005)]{Singh2005finding}
A.~Singh and A.~Moore.
\newblock Finding optimal {Bayesian} networks by dynamic programming.
\newblock Technical report, Carnegie Mellon University, 2005.

\bibitem[Spirtes and Glymour(1991)]{spirtes1991algorithm}
P.~Spirtes and C.~Glymour.
\newblock An algorithm for fast recovery of sparse causal graphs.
\newblock \emph{Social Science Computer Review}, 9\penalty0 (1):\penalty0
  62--72, 1991.

\bibitem[Spirtes and Zhang(2018)]{Spirtes2018search}
P.~Spirtes and K.~Zhang.
\newblock Search for causal models.
\newblock In M.~Maathuis, M.~Drton, S.~Lauritzen, and M.~Wainwright, editors,
  \emph{Handbook of Graphical Models}, chapter~18. CRC Press, Inc., 2018.

\bibitem[Spirtes et~al.(2000)Spirtes, Glymour, and
  Scheines]{spirtes2000causation}
P.~Spirtes, C.~N. Glymour, and R.~Scheines.
\newblock \emph{Causation, Prediction, and Search}.
\newblock MIT Press, second edition, 2000.

\bibitem[Teyssier and Koller(2005)]{teyssier2012ordering}
M.~Teyssier and D.~Koller.
\newblock Ordering-based search: A simple and effective algorithm for learning
  {Bayesian} networks.
\newblock In \emph{UAI}, 2005.

\bibitem[Wang et~al.(2021)Wang, Du, Zhu, Ke, Chen, Hao, and
  Wang]{ijcai2021-491}
X.~Wang, Y.~Du, S.~Zhu, L.~Ke, Z.~Chen, J.~Hao, and J.~Wang.
\newblock Ordering-based causal discovery with reinforcement learning.
\newblock In \emph{IJCAI}, 2021.

\bibitem[Williams(2004)]{Williams2004SimpleSG}
R.~J. Williams.
\newblock Simple statistical gradient-following algorithms for connectionist
  reinforcement learning.
\newblock \emph{Machine Learning}, 8:\penalty0 229--256, 2004.

\bibitem[Yu et~al.(2019)Yu, Chen, Gao, and Yu]{Yu19DAGGNN}
Y.~Yu, J.~Chen, T.~Gao, and M.~Yu.
\newblock {DAG-GNN}: {DAG} structure learning with graph neural networks.
\newblock In \emph{ICML}, 2019.

\bibitem[Zhang(2008)]{zhang2008completeness}
J.~Zhang.
\newblock On the completeness of orientation rules for causal discovery in the
  presence of latent confounders and selection bias.
\newblock \emph{Artificial Intelligence}, 172\penalty0 (16):\penalty0 1873 --
  1896, 2008.

\bibitem[Zhang and Hyv{\"a}rinen(2009)]{Zhang2009identifiability}
K.~Zhang and A.~Hyv{\"a}rinen.
\newblock On the identifiability of the post-nonlinear causal model.
\newblock In \emph{UAI}, 2009.

\bibitem[Zhang et~al.(2012)Zhang, Peters, Janzing, and
  Sch{\"o}lkopf]{Zhang2012kernel}
K.~Zhang, J.~Peters, D.~Janzing, and B.~Sch{\"o}lkopf.
\newblock Kernel-based conditional independence test and application in causal
  discovery.
\newblock In \emph{UAI}, 2012.

\bibitem[Zhang et~al.(2015)Zhang, Wang, Zhang, and
  Sch\"{o}lkopf]{Zhang2015estimation}
K.~Zhang, Z.~Wang, J.~Zhang, and B.~Sch\"{o}lkopf.
\newblock On estimation of functional causal models: General results and
  application to the post-nonlinear causal model.
\newblock \emph{ACM Trans. Intell. Syst. Technol.}, 7\penalty0 (2), 2015.
\newblock ISSN 2157-6904.

\bibitem[Zhang et~al.(2021)Zhang, Zhu, Kalander, Ng, Ye, Chen, and
  Pan]{gcastle}
K.~Zhang, S.~Zhu, M.~Kalander, I.~Ng, J.~Ye, Z.~Chen, and L.~Pan.
\newblock {gCastle}: {A} {Python} toolbox for causal discovery.
\newblock \emph{arXiv preprint arXiv:2111.15155}, 2021.

\bibitem[Zheng et~al.(2018)Zheng, Aragam, Ravikumar, and Xing]{zheng2018dags}
X.~Zheng, B.~Aragam, P.~Ravikumar, and E.~P. Xing.
\newblock {DAGs with NO TEARS: Continuous optimization for structure learning}.
\newblock In \emph{NeurIPS}, 2018.

\bibitem[Zheng et~al.(2020)Zheng, Dan, Aragam, Ravikumar, and
  Xing]{Zheng2020learning}
X.~Zheng, C.~Dan, B.~Aragam, P.~Ravikumar, and E.~P. Xing.
\newblock Learning sparse nonparametric {DAGs}.
\newblock In \emph{AISTATS}, 2020.

\bibitem[Zhu et~al.(2020)Zhu, Ng, and Chen]{Zhu2020causal}
S.~Zhu, I.~Ng, and Z.~Chen.
\newblock Causal discovery with reinforcement learning.
\newblock In \emph{ICLR}, 2020.

\end{thebibliography}

\newpage
\begin{appendices}
\section{Proofs}
\subsection{Proof of Lemma~\ref{lemma1}}
\label{prooflemma1}
\begin{proof}
It suffices to show that if $X_j$ is not a parent of  $X_i$ in $\cal H$, then $X_j$ is not a parent of $X_i$ in $\widetilde{\cal G}$, either. That $X_j$ is not a parent of $X_i$ in $\cal H$ indicates $B_{ji}=0$. Therefore, $h_i(B_i\circ X)$ is a constant function w.r.t.~$X_j$. For the reduced SEM with functions $\tilde f_i$'s and causal DAG $\widetilde{\cal G}$, we conclude that $X_j\notin X_{\widetilde{\mathrm{pa}}(i)}$ and the input arguments of $\tilde{f}_i$ do not contain $X_j$. Thus, $X_j$ cannot be a parent of $X_i$ in $\widetilde{\cal G}$.
\end{proof}

\subsection{Proof of Proposition~\ref{prop:identification}}
\label{proofproposition1}
\begin{proof}
Recall that the reduced SEM with  $\tilde f_i$'s and  graph $\widetilde{\cal G}$ satisfies the causal minimality condition and has the same distribution $P(X)$. With the identifiability result of restricted ANMs \citep[Theorem~28]{Peters2014causal}, we know that $\widetilde{\cal G}$ is identical to $\cal G$. Applying Lemma~\ref{lemma1} completes the proof.
\end{proof}

\section{Derivation of Logistic Distribution}
\label{sec:gumbel_logistic}
We provide a derivation that if two independent variables $X, Y\sim \operatorname{Gumbel}(0,1)$, then $Z=X-Y\sim \operatorname{Logistic}(0,1)$. We will show that the CDF of $Z$ matches the CDF of $\operatorname{Logistic}(0,1)$.

The CDF of $X$ is given by
\[\operatorname{Pr}\,(X\leq x) = e^{-e^{-x}},\quad x\in\mathbb R.\]
We then get 
\[\operatorname{Pr}\left(e^X\leq x'\right) = \operatorname{Pr}\left(X\leq\log x'\right)=e^{-x'}, \quad x'\in\mathbb R^+,\]
which indicates $e^X$ follows the exponential distribution with the rate parameter being $1$.

To find the distribution of $Z$, we notice that $Z=\log (e^{X-Y})$. Then for $z\in\mathbb R$, we have
\begin{align}
\operatorname{Pr}\,(Z\leq z) 
& = \operatorname{Pr}\left(\log\left(e^{X-Y}\right)\leq z\right)\nonumber\\
& =\operatorname{Pr}\left(e^X/e^Y\leq e^z\right) \nonumber\\
& = \int_{x'=0}^\infty\int_{y'=x'/e^{z}}^\infty e^{-x'} e^{-y'} dx' dy'\nonumber\\
& = \frac{1}{1+e^{-z}},\nonumber
\end{align}
which is exactly the CDF of $\operatorname{Logistic}(0,1)$.

\section{Further Details on Optimization with Augmented Lagrangian Method}
\label{sec:optimization}

\subsection{Parameter Choices}\label{sec:optimization_parameter_choice} 
As shown by \citet{jang2017categorical}, the Gumbel-Softmax approach is effective for single sample gradient estimation, so we simply pick $b=1$ which is found to work well in our experiments. We pick  $\gamma=0.25$, same as in NOTEARS and DAG-GNN, and initialize the estimate of Lagrange multiplier $\alpha$ to $0$, i.e., $\alpha^0=0$. We set the initial value of $\rho$ by $\rho^0=10^{-\left \lceil 0.3d \right \rceil}$ for $d$-node graphs, with minor fine tuning. This is because we initialize the logits $U_{ji}$'s to be $0$ and consequently $s(U^0)$ would be very large  with high probability for large graphs. Since  both $\alpha$ and $\rho$ do not decrease during training, a large $\rho^0$ may make the optimization somewhat `omit' the score function and the resulting graph would be a DAG with high score. However, a  small initial value $\rho^0$ would result in a long training time, so we  choose larger $\beta$ for larger graphs to accelerate training. In particular, we find that $\beta=5$, $15$, $300$ and $8000$ work well for graphs with $10$, $20$, $50$ and $100$ nodes, respectively. These choices are found on synthetic datasets with causal additive model and known true causal graphs (see  Appendix~\ref{sec:hyperparameters} for further discussions). For graphs with other sizes, a linear interpolation on the logarithm scale of the graph sizes can be used as the choice for $\beta$, and one may also consider fine tuning using some synthetic datasets with known true graphs. We use the Adam optimizer \citep{Kingma2014adam} with a learning rate of $3\times 10^{-2}$ and $1000$ training iterations to approximately solve the subproblem in Eq.~(\ref{eqn:al_2}). Note that for $100$-node graphs, we set the number of training iterations to $2500$ as it may take longer to converge on larger graphs.

\subsection{Stopping Criterion} Simply using a single sample estimate $s(U^t)$ for stopping criterion may make the optimization stop too early. We consider further steps: 
\begin{itemize}
	\item we choose the tolerance $\xi$ to be small to lower the probability of stopping the algorithm early. Particularly, we pick $\xi=10^{-10}$ in our experiment which is much smaller than that used in NOTEARS and GraN-DAG. 
	\item we set another stopping criterion $\operatorname{tr}\big(e^{\sigma(U^{t}/\tau)}\big)<\xi$ which uses only the logits. This criterion will be satisfied only when the entries $\sigma(U^{t}_{ji}/\tau)$, where $i,j$ are such that the edge $X_j\to X_i\notin E$ for some DAG with $E$ as edge set, are nearly zeros. If an edge indeed helps minimize the score function,  then the logit will be pushed to a relatively large positive value so that the score function is minimized in the expected sense. Thus, as a byproduct, we can then readily use $\sigma(U^{t}/\tau)$ as our learned matrix which indicates a DAG after a hard thresholding at threshold $0.5$.
	\item notice that $s(U^{t})$  and $\operatorname{tr}\left(e^{\sigma(U^{t}/\tau)}\right)$ are evaluated at the end of  step $t$ of the augmented Lagrangian method. This also helps avoid terminating the optimization too early.
\end{itemize}

\begin{figure}[ht]
\centering
\subfloat[$s(U)$ and $\operatorname{tr}(e^{\sigma(U/\tau)})$.]{
\label{fig:acyclic} 
  \includegraphics[width=0.66\linewidth]{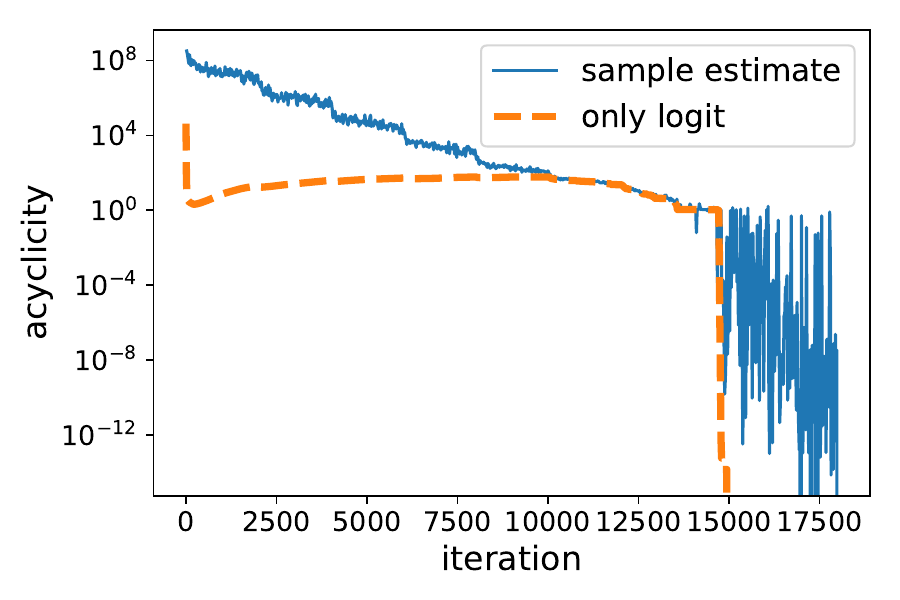}
}\\
\subfloat[Augmented Lagrangian.]{
\label{fig:overfit}
  \includegraphics[width=0.66\linewidth]{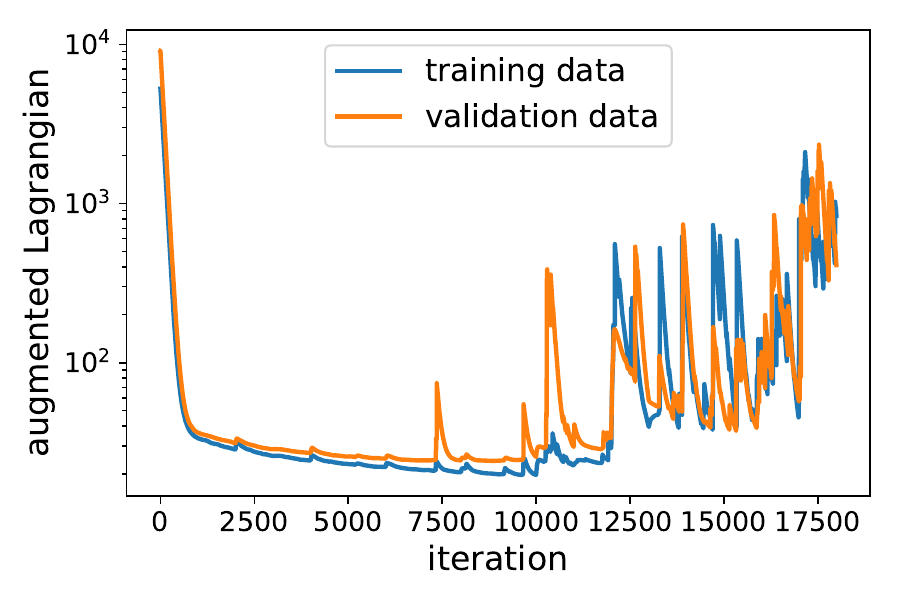}
}
\caption{Training trajectories.} 
\label{fig:training_trajectories}
\end{figure}

In Figure~\ref{fig:acyclic}, we provide an empirical study regarding $s(U)$ and $\operatorname{tr}\left(e^{\sigma(U/\tau)}\right)$, on the GP dataset with $50$-node ER1 graph. Figure~\ref{fig:acyclic} and similar results on other datasets validate the effectiveness of the use of both stopping criteria. As discussed in Section~\ref{mask_acyclicity}, the matrix $\sigma(U^{t}/\tau)$ at the last step will  be further processed to output our inferred causal graph.

\subsection{Overfitting}\label{sec:overfit}
Unlike DAG-GNN and GraN-DAG that require a held-out dataset to avoid overfitting, we use the whole dataset during training. To investigate whether MCSL-MLP incurs overfitting, we generate a validation dataset of $3000$ samples and monitor the augmented Lagrangian on both training and validation datasets. Here the validation dataset is not used for structure learning. A typical example on a GP dataset with $50$-node ER1 graph is plotted in Figure~\ref{fig:training_trajectories}.

We observe two phases: 1) the augmented Lagrangian is optimized to decrease on both  datasets  in the first $10000$ iterations; and 2) with increased estimate of Lagrange multiplier and penalty parameter, the acyclicity term has a larger effect on the augmented Lagrangian, making the acyclicity term decrease towards the predefined tolerance and the least squares losses tend to increase. The augmented Lagrangian also oscillates in the second phase, due to the varying logits $U_{ji}$'s and the increasing estimate of Lagrange multiplier and penalty parameter. The similar behaviors in both phases indicate that MCSL-MLP can use all the observed data for training and does not need a held-out dataset for validation, at least in our experiments.

\section{Further Experiment Details and Results} \label{sec:supplementary_experiments}
\subsection{Hyperparameters} \label{sec:hyperparameters}
MCSL-MLP uses $4$-layer MLPs with $16$ Leaky ReLU hidden units as the model functions, where the weights are initialized using the Xavier uniform initialization \citep{Glorot2010TrainingDNN}. Note that if the logit $U_{ji}$ in $\mathsf g_\tau(U)$ is small (e.g., $-2$ or $-3$), then the Gumbel-Sigmoid output is highly likely to be deactivated with a small temperature, thus increasing the difficulty of gradient-based optimization. We therefore initialize each logit $U_{ji}$ to $0$. For Gumbel-Sigmoid, we set a fixed temperature $\tau=0.2$ which is found to work well. Finally, the $\ell_1$ penalty weight is chosen as $2\times 10^{-3}$ to slightly control false discoveries, followed by a pruning step to further remove spurious edges. The hyperparameters related to the augmented Lagrangian method have been discussed in Appendix~\ref{sec:optimization_parameter_choice}. 

In practice, however, one could not perform hyperparameter tuning directly on the observed data as the ground truth is not available. Similar to GraN-DAG, we conduct experiments on synthetic data with known causal graphs to search for the hyperparameters, and then use these hyperparameters for all the experiments. In particular, we choose a causal additive data model and the CAM algorithm \citep{Buhlmann2014cam}, which is specifically designed for this type of data models, can serve as a good benchmark for our method.

\begin{table*}[t]
\centering
\caption{Empirical results on nonlinear SEMs with Gaussian processes.}
\vspace{-0.5em}
\label{tab:additional_results_gp}{\notsotiny
\begin{tabular}{lll|ll|ll|llllll}
\toprule
~& \multicolumn{2}{l}{ER1 with $20$ nodes} & \multicolumn{2}{l}{ER4 with $20$ nodes} 
    & \multicolumn{2}{l}{ER1 with $100$ nodes} & \multicolumn{2}{l}{ER4 with $100$ nodes}\\
    \cmidrule(lr){2-3} \cmidrule(lr){4-5} \cmidrule(lr){6-7} \cmidrule(lr){8-9}
& \multicolumn{1}{l}{SHD} & \multicolumn{1}{l}{TPR} & \multicolumn{1}{l}{SHD} & \multicolumn{1}{l}{TPR}
    & \multicolumn{1}{l}{SHD} & \multicolumn{1}{l}{TPR} & \multicolumn{1}{l}{SHD} & \multicolumn{1}{l}{TPR}\\ \midrule
MCSL-MLP   &    {\bf 1.4\,$\pm$\,1.1} & {\bf 0.93\,$\pm$\,0.04} &  {\bf 15.8\,$\pm$\,2.4}  & {\bf 0.80\,$\pm$\,0.03}
    &  27.6\,$\pm$\,12.3 & 0.79\,$\pm$\,0.09 & {\bf 121.2\,$\pm$\,11.9} & {\bf 0.76\,$\pm$\,0.03}\\
GraN-DAG   &    {5.7\,$\pm$\,2.1} & {0.79\,$\pm$\,0.04} & {\bf 28.4\,$\pm$\,4.9}  & {\bf 0.69\,$\pm$\,0.04}
    &   {\bf 18.6\,$\pm$\,5.8} & {\bf 0.83\,$\pm$\,0.06} &  {\bf 156.4\,$\pm$\,12.2} & {\bf 0.67\,$\pm$\,0.04}\\
CAM        &    {\bf 1.0\,$\pm$\,2.0} & {\bf 0.97\,$\pm$\,0.06} &  {\bf 16.4\,$\pm$\,1.5}  & {\bf 0.81\,$\pm$\,0.02}
    &   {\bf 6.4\,$\pm$\,2.6} & {\bf 0.97\,$\pm$\,0.02} &  {\bf 77.2\,$\pm$\,22.4} & {\bf 0.83\,$\pm$\,0.04}\\
NOTEARS-MLP~&   {\bf 0.8\,$\pm$\,1.3} & {\bf 0.96\,$\pm$\,0.07}  & 40.4\,$\pm$\,7.2  & 0.49\,$\pm$\,0.07
    &  {\bf 11.2\,$\pm$\,4.5} & {\bf 0.89\,$\pm$\,0.04} & 216.6\,$\pm$\,18.3 & 0.46\,$\pm$\,0.04\\
    
SAM     &    10.8\,$\pm$\,2.3  & 0.53\,$\pm$\,0.11   & 69.2\,$\pm$\,3.3  &  0.12\,$\pm$\,0.03
    &   N/A  & N/A & N/A & N/A\\
DAG-GNN    &    6.0\,$\pm$\,0.89 & 0.67\,$\pm$\,0.05 & 72.2\,$\pm$\,6.5  & 0.09\,$\pm$\,0.05
    &  46.8\,$\pm$\,8.9 & 0.49\,$\pm$\,0.06 & 375.2\,$\pm$\,21.2 & 0.07\,$\pm$\,0.02\\
NOTEARS    &    6.8\,$\pm$\,3.8 & 0.75\,$\pm$\,0.08 & 69.0\,$\pm$\,6.1  & 0.14\,$\pm$\,0.04
    &  35.4\,$\pm$\,9.4 & 0.67\,$\pm$\,0.05 & 346.8\,$\pm$\,21.2 & 0.15\,$\pm$\,0.02\\
GES        &    5.0\,$\pm$\,2.9 & 0.82\,$\pm$\,0.08 & 58.2\,$\pm$\,3.3  & 0.28\,$\pm$\,0.03
    &  24.4\,$\pm$\,9.1 & 0.78\,$\pm$\,0.07 & 289.6\,$\pm$\,21.7 & 0.30\,$\pm$\,0.03\\
PC         &    5.4\,$\pm$\,2.4 & 0.86\,$\pm$\,0.09 & 47.6\,$\pm$\,6.0  & 0.41\,$\pm$\,0.06
    &  40.2\,$\pm$\,6.7 & 0.74\,$\pm$\,0.05 & 228.2\,$\pm$\,21.1 & 0.46\,$\pm$\,0.03\\
GES-GS     &  4.4\,$\pm$\,1.9 & 0.83\,$\pm$\,0.10   &  57.0\,$\pm$\,4.1   &  0.32\,$\pm$\,0.03
    &  N/A  & N/A & N/A & N/A\\
PC-KCI     &    11.58\,$\pm$\,3.04  & 0.83\,$\pm$\,0.12   & N/A\,$\pm$\,N/A  &  N/A\,$\pm$\,N/A
    &  N/A & N/A & N/A & N/A \\
\bottomrule
\end{tabular}}
\end{table*}

\begin{table*}[htbp]
\centering
\caption{Empirical results on nonlinear SEMs with quadratic functions.}
\vspace{-0.5em}
\label{tab:results_quadratic}{\notsotiny
\begin{tabular}{lll|ll|ll}
\toprule
~& \multicolumn{2}{l}{$10$ nodes} & \multicolumn{2}{l}{$20$ nodes} & \multicolumn{2}{l}{$50$ nodes}\\
    \cmidrule(lr){2-3} \cmidrule(lr){4-5} \cmidrule(lr){6-7}
                 & \multicolumn{1}{l}{SHD} & \multicolumn{1}{l}{TPR} & \multicolumn{1}{l}{SHD} & \multicolumn{1}{l}{TPR} & \multicolumn{1}{l}{SHD} & \multicolumn{1}{l}{TPR}\\ \midrule
MCSL-MLP         &  {\bf 4.0\,$\pm$\,2.9}  & {\bf 0.65\,$\pm$\,0.26} &  {\bf 4.6\,$\pm$\,3.6}  & {\bf 0.80\,$\pm$\,0.14} & {\bf 19.2\,$\pm$\,7.3} & {\bf 0.70\,$\pm$\,0.07} \\
MCSL-quad         &  {\bf 3.4\,$\pm$\,2.8}  & {\bf 0.73\,$\pm$\,0.20} &  {\bf 4.4\,$\pm$\,3.0}  & {\bf 0.80\,$\pm$\,0.13} & {\bf 20.4\,$\pm$\,8.7} & {\bf 0.66\,$\pm$\,0.11} \\
GraN-DAG         &       6.0\,$\pm$\, 3.3  &      0.59\,$\pm$\,0.21  &       8.2\,$\pm$\,3.0   &      0.78\,$\pm$\,0.19  &      49.0\,$\pm$\,12.3  &      0.53\,$\pm$\,0.11 \\
CAM              &       5.4\,$\pm$\, 3.2  &      0.64\,$\pm$\,0.20  &      11.8\,$\pm$\,4.2   &      0.69\,$\pm$\,0.18  &      32.6\,$\pm$\,11.5  &      0.50\,$\pm$\,0.13 \\
NOTEARS-MLP      &       7.6\,$\pm$\, 4.7  &       0.42\,$\pm$\,0.32  &      11.8\,$\pm$\,5.5   &      0.49\,$\pm$\,0.22  &      39.8\,$\pm$\,11.1  &      0.34\,$\pm$\,0.14 \\
NOTEARS-quad~~   &      9.8\,$\pm$\,3.2  &      0.22\,$\pm$\,0.08  &       13.4\,$\pm$\,4.5  &      0.40\,$\pm$\,0.12  &      46.0\,$\pm$\,7.0   &      0.23\,$\pm$\,0.05 \\
\bottomrule
\end{tabular}}
\end{table*}

\subsection{Results for Nonlinear SEMs with Gaussian Processes} 
\label{more_exp_gp}
This section  provides additional results for the nonlinear SEMs with Gaussian processes in Section~\ref{exp_gp}, as shown in Table~\ref{tab:additional_results_gp}. Here N/A means that the corresponding experiments could not finish within $12$ hours; since the corresponding methods have been shown to perform poorly in other settings, we simply stopped them.

\subsection{Experiment Setup and Results for Nonlinear SEMs with Quadratic Functions}
\label{more_exp_quadratic}
This section provides further experiment details and results for the nonlinear SEMs with quadratic functions in Section~\ref{exp_quadratic}.

\paragraph{Setup} Our setup here is similar to \citet{Zhu2020causal}.
For each variable $X_i$, we expand its set of parental nodes $X_{\mathrm{pa}(i)}$ to obtain both first- and second-order feature terms. That is, for $X_{\mathrm{pa}(i)}=\{X_{j_1}, X_{j_2},\ldots, X_{j_{d'}}\}$ with $d'$ being the cardinality of $X_{\mathrm{pa}(i)}$, the first-order feature terms correspond to $X_{j_1},\ldots, X_{j_{d'}}$, and the second-order feature terms are $X_{j_1}^2,\ldots, X_{j_{d'}}^2, X_{j_1}X_{j_2},\ldots, X_{j_1}X_{j_{d'}}, X_{j_2}X_{j_3},\ldots, X_{j_{d'-1}}X_{j_{d'}}$. We set the coefficient of each feature term to be either $0$ or sampled uniformly from $[-1, -0.5]\cup[0.5,1]$, with equal probability. Moreover, if a parental variable is not contained in any of the feature terms with a non-zero coefficient, the corresponding edge is removed from the DAG. We consider ER1 graphs with $d\in\{10,20,50\}$ nodes. Other settings such as noise distribution and number of samples are the same as Section \ref{exp_gp}.

For this data model, there may exist very large variable values that would lead to numerical issues for the followed gradient-based structure learning methods including NOTEATS, GraN-DAG, and MCSL-MLP. We therefore limit the diameter of each underlying causal graph to $3$ and normalize each variable's values (before incorporating the corresponding additive noises) by dividing the total number of corresponding first- and second-order terms. The normalized data are then used as the observed data in the experiment.

\paragraph{Quadratic regression based pruning method}  \citet{Zhu2020causal} also adopted a Quadratic Regression based pruning method (QR pruning) to remove spurious edges. QR pruning applies quadratic regression to each variable $X_i$ and its parents $X_{\mathrm{pa}(i)}$  indicated from the estimated DAG, followed by thresholding on the resulting coefficients of both first- and second-order terms. In this experiment, we pick a threshold $0.1$. If the coefficient of an interaction term, e.g., $X_{i_1}X_{i_2}$, is non-zero after thresholding, then we have two directed edges which are $X_{i_1}\to X_i$ and $X_{i_2}\to X_i$.

\paragraph{Empirical results} 
The empirical results  are presented in Table~\ref{tab:results_quadratic}.

\begin{table*}[htbp]
\centering
\caption{Empirical results on post-nonlinear causal models.}
\vspace{-0.5em}
\label{tab:results_pnl}{\notsotiny
\begin{tabular}{lll|ll|ll}
\toprule
~& \multicolumn{2}{l}{$10$ nodes} & \multicolumn{2}{l}{$20$ nodes} & \multicolumn{2}{l}{$50$ nodes}\\
    \cmidrule(lr){2-3} \cmidrule(lr){4-5} \cmidrule(lr){6-7}
                 & \multicolumn{1}{l}{SHD} & \multicolumn{1}{l}{TPR} & \multicolumn{1}{l}{SHD} & \multicolumn{1}{l}{TPR} & \multicolumn{1}{l}{SHD} & \multicolumn{1}{l}{TPR}\\ \midrule
MCSL-MLP         &  {\bf 9.4\,$\pm$\,6.5}  & {\bf 0.85\,$\pm$\,0.12} & {\bf 13.2\,$\pm$\,2.7} &  {\bf 0.80\,$\pm$\,0.06} & {\bf 70.0\,$\pm$\,22.3} & {\bf 0.58\,$\pm$\,0.13} \\
GraN-DAG~~       &      16.6\,$\pm$\,5.4   &      0.03\,$\pm$\,0.07  &      24.8\,$\pm$\,4.0  &       0.10\,$\pm$\,0.05  &      84.0\,$\pm$\,11.1  &      0.15\,$\pm$\,0.06  \\
CAM              &      22.0\,$\pm$\,8.0   &      0.12\,$\pm$\,0.08  &      35.0\,$\pm$\,8.8  &       0.21\,$\pm$\,0.09  &     108.2\,$\pm$\,4.7   &      0.26\,$\pm$\,0.11 \\
\bottomrule
\end{tabular}}
\end{table*}

\subsection{Results for Post-Nonlinear Causal Models} 
\label{sec:more_exp_pnl}
This section  provides additional results for the post-nonlinear models in Section~\ref{sec:exp_pnl}, as shown in Table~\ref{tab:results_pnl}.

\subsection{Scale-Free Graphs}
\label{sec:scale_free_exp}
We conduct an empirical study with $50$-node scale-free graphs, by considering the GP dataset from Section~\ref{exp_gp}. CAM performs the best with SHDs $12.6\pm 8.0$ and $78.4\pm 9.1$ for graphs with $50$ and $200$ edges on average, respectively, while GraN-DAG has SHDs $14.8\pm 8.4$ and $126.8\pm 4.8$, respectively. Our method is still competitive to the best method CAM with SHDs $15.0\pm 8.2$ and $82.0\pm 4.7$.

\subsection{Different Sample Sizes}
\label{sec:consistency}
\begin{figure}[ht]
\centering
\subfloat[SHD.]{
  \includegraphics[width=0.66\linewidth]{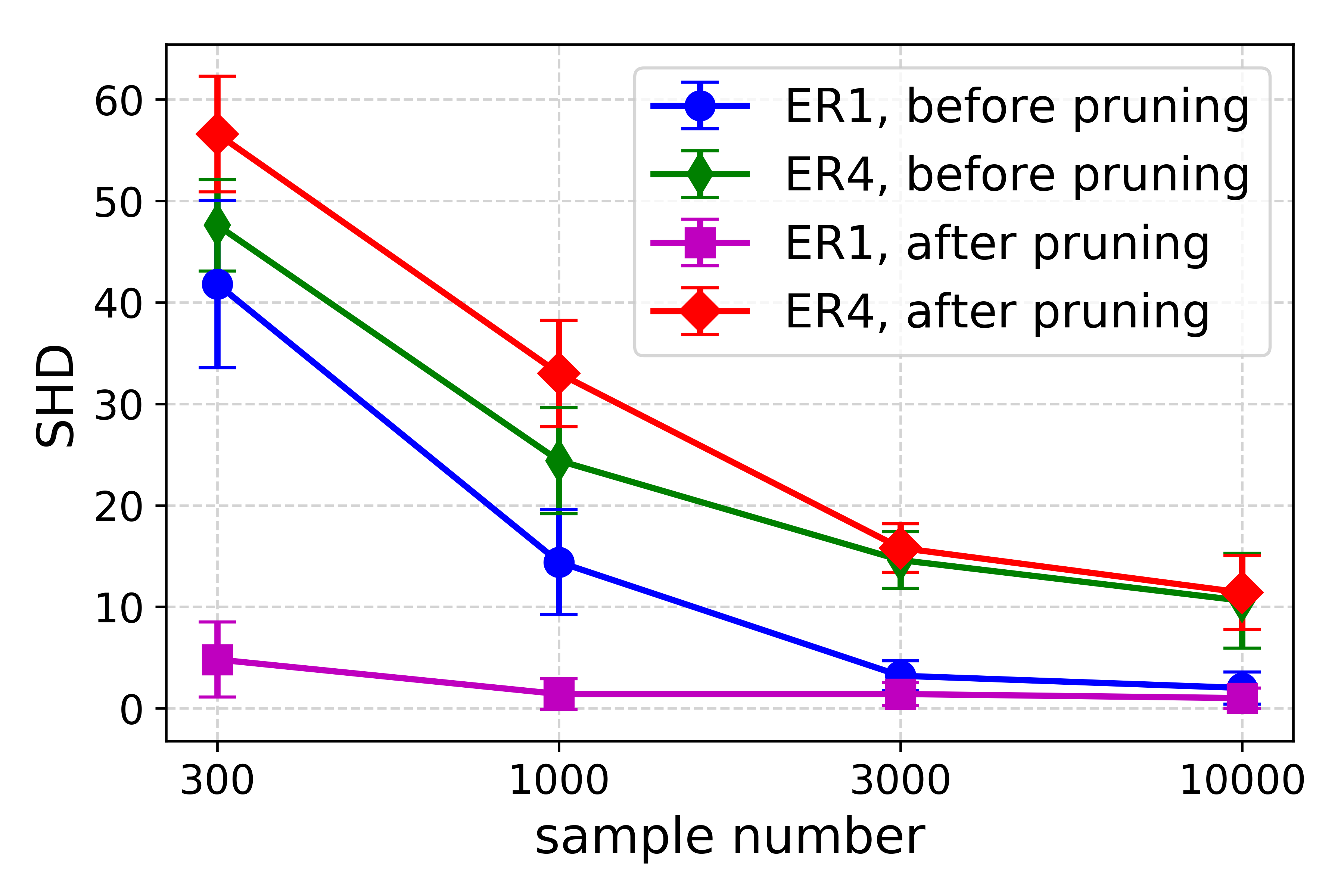}
}\\
\subfloat[TPR.]{
  \includegraphics[width=0.66\linewidth]{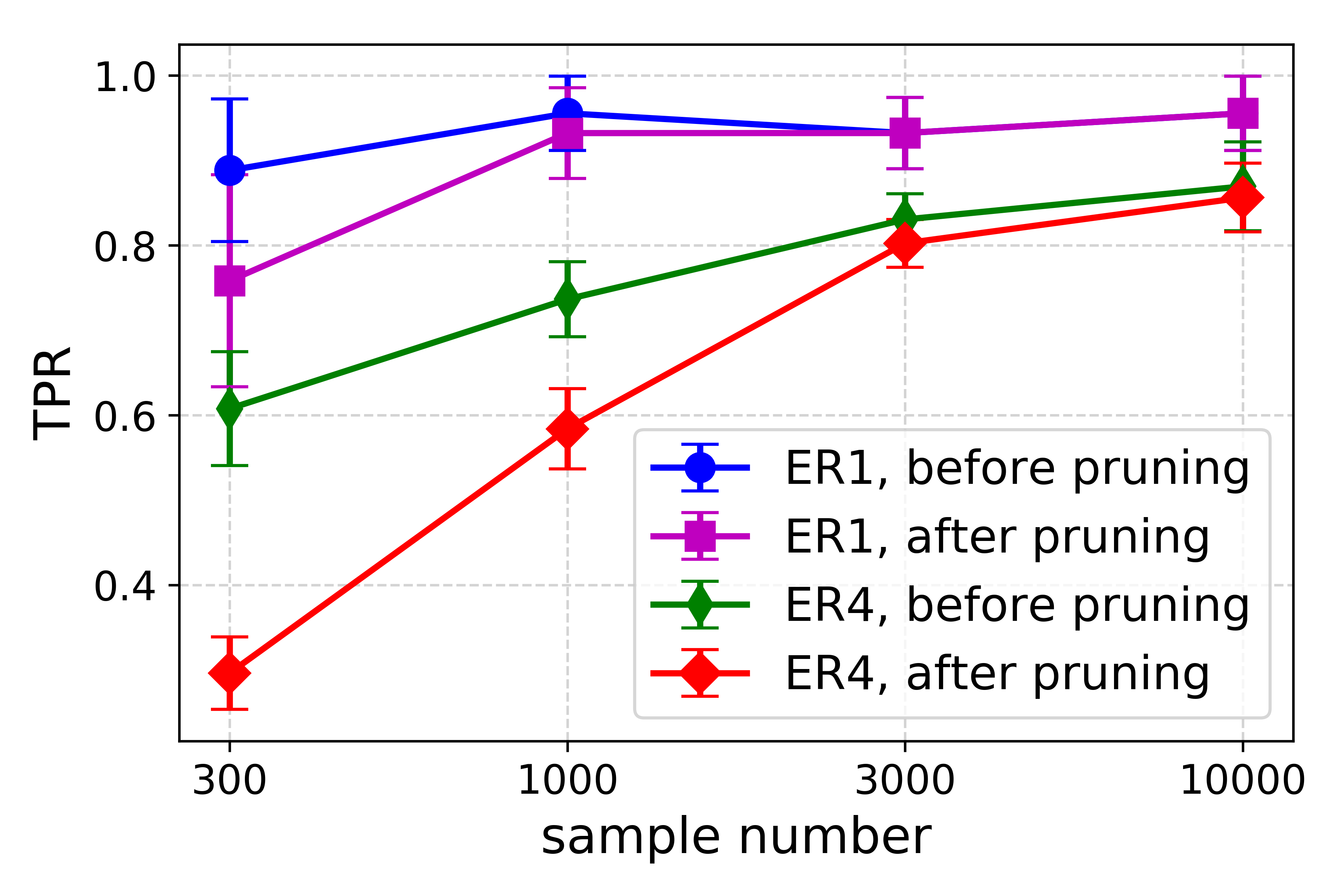}
}
\caption{Different sample sizes.} 
\label{fig:sample_size}
\end{figure}

We also conduct an empirical analysis of MCSL-MLP with different sample numbers. We pick the GP dataset with $20$ nodes and $n=\{300, 1000, 3000, 10000\}$ samples. As shown in Figure~\ref{fig:sample_size}, a larger number of samples generally leads to a better performance. Another interesting observation from ER4 graphs is that the CAM pruning method seems to remove more true positives than false discoveries and somewhat degrades the performance, when the true causal graphs are dense.

\subsection{Sigmoid vs.~Gumbel-Sigmoid}
\label{sec:sigmoidvsgumbelsigmoid}
\begin{figure}[!h]
\vspace{-1.0em}
\centering
\subfloat[Sigmoid.]{
  \includegraphics[width=0.44\linewidth]{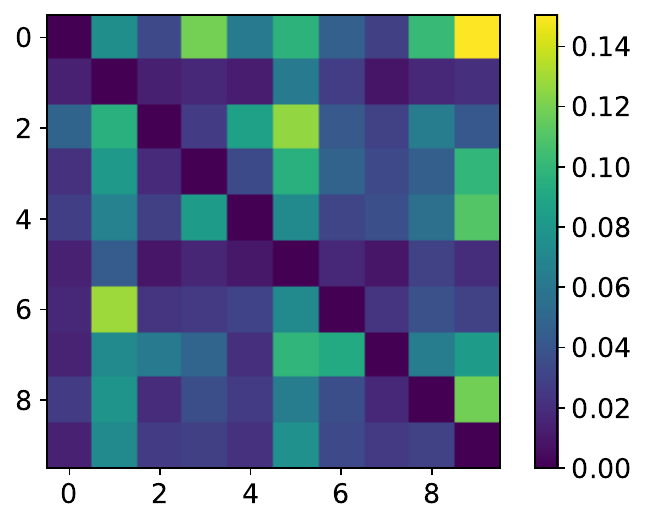}
\label{fig:visualize_sigmoid_graph}
}
\subfloat[Gumbel-Sigmoid.]{
  \includegraphics[width=0.44\linewidth]{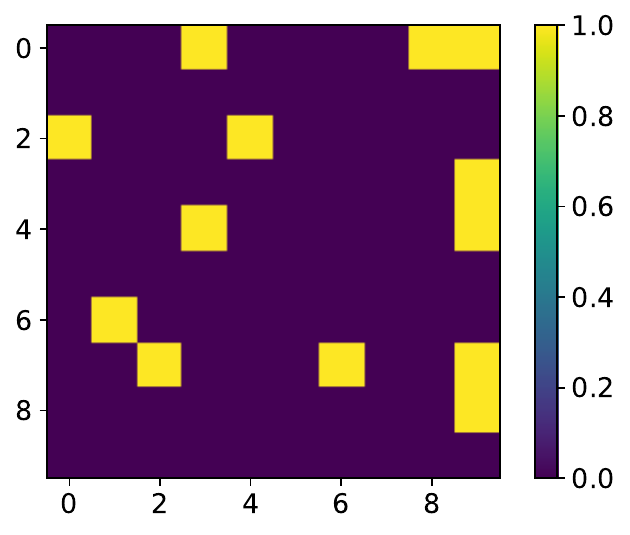}
  \label{fig:visualize_gumbel_sigmoid_graph}
}
\caption{Visualization of the estimated matrices.}  
\label{fig:visualize_graphs}
\vspace{-0.5em}
\end{figure}

We provide an example to show the effectiveness of applying Gumbel-Sigmoid to approximating the binary adjacency matrix. We consider a dataset with $10$-node ER1 graph from the experiment in Section~\ref{exp_gp}. Notice that this dataset is only for illustration purpose and similar results are found with other datasets as well. We apply logistic sigmoid and Gumbel-Sigmoid with MCSL-MLP and visualize the resulting adjacency matrices in Figure~\ref{fig:visualize_graphs}. 

Figure~\ref{fig:visualize_sigmoid_graph} shows that the estimated entries from logistic sigmoid functions lie in a small range near $0$. This is because the acyclicity and $\ell_1$ penalty terms can be made very small if all the entries are close to $0$. Consequently, it becomes hard to choose a proper threshold to identify the edges from the estimate. Gumbel-Sigmoid alleviates this problem by enforcing the estimated entries to be close to either $0$ or $1$, as shown in Figure~\ref{fig:visualize_gumbel_sigmoid_graph}. We can then easily identify the positive edges by a threshold at $0.5$. For this example, we indeed recover the true graph.

\subsection{Pruning} \label{exp_pruning}
We investigate the effect of the additional pruning step on the proposed method. A useful approach is the CAM pruning \citet{Buhlmann2014cam} that applies significance testing of covariates using generalized additive models and declares significance if the $p$-values do not exceed a predefined value. Though the true causal relationships may not follow causal additive assumption, the CAM pruning usually performs well.

We use the GP dataset from the experiment in Section~\ref{exp_gp} and compare MCSL-MLP with or without pruning on ER1 and ER4 graphs with $d\in\{10,20,50, 100\}$ nodes. Notice that this pruning step is necessary for CAM and GraN-DAG. CAM first estimates a topological order of the variables and the graph has all the possible edges that does not violate the acyclicity constraint. It then uses pruning to remove spurious edges. GraN-DAG estimates an equivalent adjacency matrix and then removes an edge if the corresponding entry has the smallest value in the absolute Jacobian matrix until a DAG is obtained. Note that this DAG typically contains many spurious edges and pruning is needed to reduce false discoveries, as shown by the ablation study in \citet{Lachapelle2019grandag}.

\begin{figure*}[htbp]
\centering
\subfloat[ER1 graphs.]{
  \includegraphics[width=0.92\textwidth]{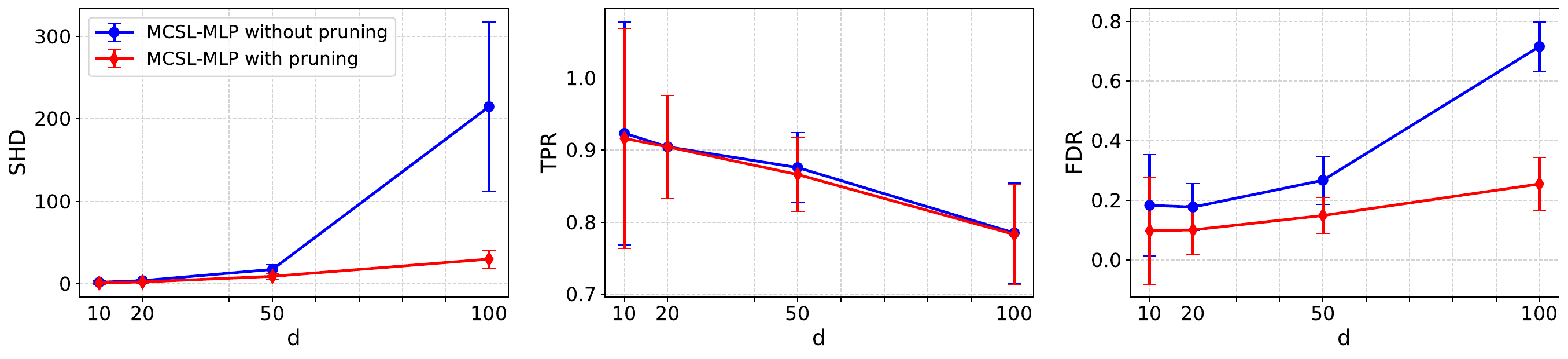}
  \label{fig:pruning_result_deg2}
}\\
\subfloat[ER4 graphs.]{
  \includegraphics[width=0.92\textwidth]{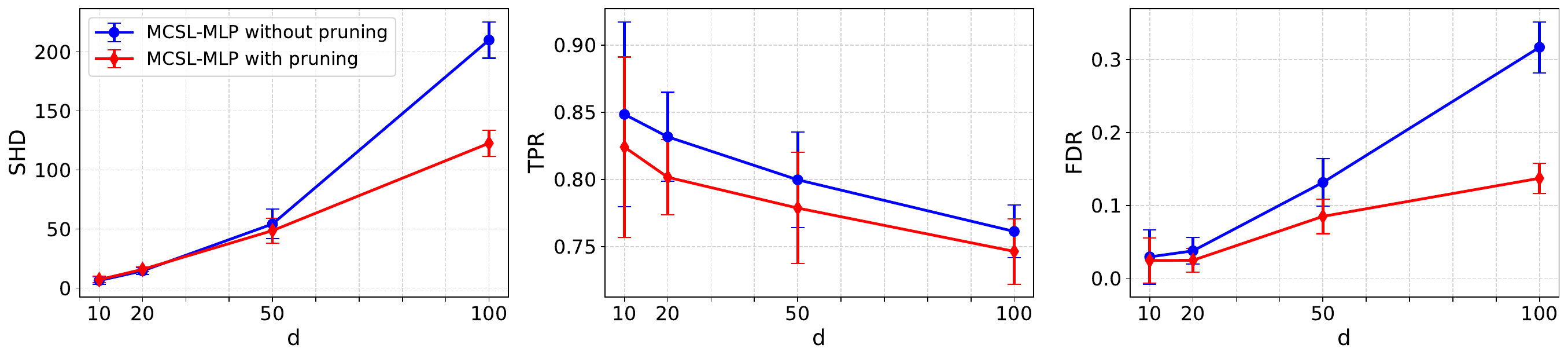}
  \label{fig:pruning_result_deg8}
}
\caption{The effect of pruning on MCSL-MLP.}  
\label{fig:pruning_result}  
\end{figure*}

Figure~\ref{fig:pruning_result} reports the empirical results in terms of SHD, TPR and also False Discovery Rate (FDR). We observe that the additional pruning step reduces the SHD and FDR much, and has little effect on the TPR on ER1 graphs. For ER4 graphs that are denser, CAM pruning reduces both FDR and TPR. Nevertheless, the overall metric SHD is improved by CAM pruning, and the difference grows as the graph size increases. This experiment demonstrates the practical importance of applying an additional pruning step for MCSL-MLP.

\subsection{Computational Complexity and Training Time}
\label{sec:complexity}
Similar to NOTEARS and GraN-DAG, MCSL-MLP requires evaluations of matrix exponential with $\mathcal O(d^3)$ cost per iteration. To reduce the number of $\mathcal O(d^3)$ iterations required to converge, NOTEARS adopts the proximal quasi-Newton algorithm. Although GraN-DAG uses a gradient-based method like ours, it has been observed that GraN-DAG performs fewer iterations than NOTEARS in practice and that the evaluation of matrix exponential does not dominate the total cost at each iteration for graphs with $100$ nodes or less \citep{Lachapelle2019grandag}. 

\begin{figure}[!t]
  \begin{center}
    \includegraphics[width=0.66\linewidth]{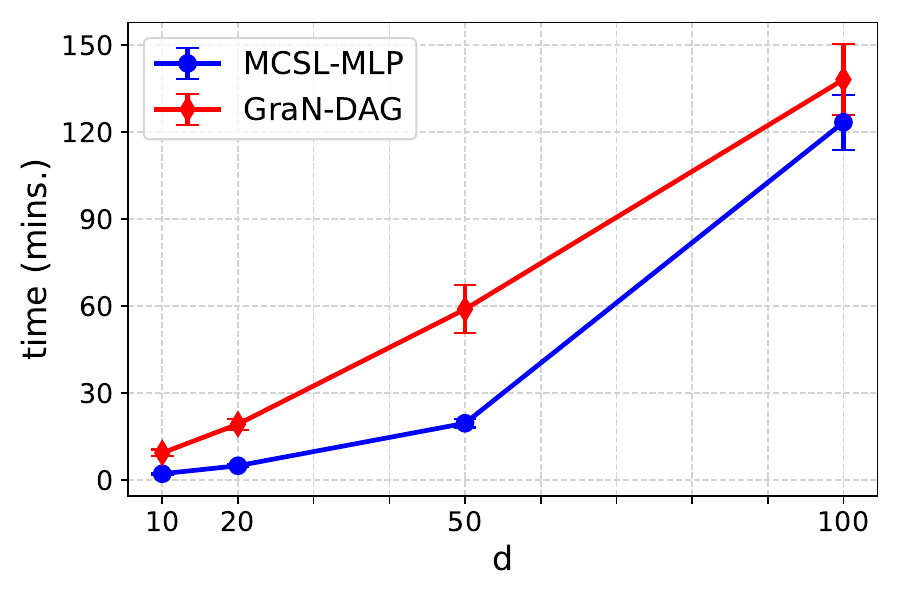}
  \end{center}
  \caption{Training time.}
  \label{fig:training_time}
\end{figure}

To validate whether MCSL-MLP behaves similarly, we simply compare the training time of MCSL-MLP with GraN-DAG using the GP dataset from Section \ref{exp_gp}, as shown in Figure~\ref{fig:training_time}. The experiments are run on a standard NC6 instance on Azure cloud, with 6-core Intel Xeon 2.6GHz CPU and one-half Nvidia Tesla K80 GPU. MCSL-MLP takes a shorter time than GraN-DAG across all graph sizes, showing that the $\mathcal{O}(d^3)$ evaluation costs of matrix exponential are not a problem to MCSL-MLP, at least for problems with $100$ nodes or less.  There is an increase in the training time of MCSL-MLP on $100$-node graphs, as we use an increased number of training iterations to solve the subproblem in Eq.~\eqref{eqn:al_2}. We believe that a faster algorithm, e.g., a second-order algorithm as in \cite{Zheng2020learning}, can be adopted to further increase the efficiency of MCSL-MLP for larger problems, which is treated as a future work.

\end{appendices}

\end{document}